\documentclass[oneside]{article}

\pdfoutput=1 

\usepackage{amsmath}
\usepackage{amssymb}
\usepackage{bm}
\usepackage{tikz}
\usepackage{pgfplots}
\usepackage{array}
\usepackage{microtype}
\usepackage{subfigure}
\usepackage{xfrac}
\usepackage{fancyhdr} 

\newcommand{\x}{\bm{x}}
\newcommand{\y}{y}
\newcommand{\upref}{\bm{u}}
\newcommand{\vpref}{\bm{v}}
\newcommand{\data}{\mathcal{D}}
\newcommand{\param}{\bm{\theta}}
\newcommand{\argmax}{ \operatorname*{arg \max}} 
\newcommand{\argmin}{ \operatorname*{arg \min}} 
\newcommand{\ourmethod}{BALD}
\newcommand{\E}{\mathbb{E}}
\newcommand{\rmI}{\mathrm{I}}
\newcommand{\rmh}{\mathrm{h}}
\newcommand{\rmH}{\mathrm{H}}

\definecolor{mycolor1}{rgb}{0,0.4,0}
\definecolor{mycolor2}{rgb}{0,1,1}
\definecolor{mycolor3}{rgb}{1,0,1}
\definecolor{mycolor4}{rgb}{1,0.8,0.5}
\definecolor{mycolor5}{rgb}{0.7,0.4,0.01}

\begin{document}

\title{ Bayesian Active Learning for Classification and Preference Learning }

\author{Neil Houlsby, Ferenc Husz\'{a}r, Zoubin Ghahramani, M\'{a}t\'{e} Lengyel \\ Computational and Biological Learning Laboratory \\ University of Cambridge}

\maketitle

\begin{abstract}

Information theoretic active learning has been widely studied for probabilistic models. For simple regression an optimal myopic policy is easily tractable. However, for other tasks and with more complex models, such as classification with nonparametric models, the optimal solution is harder to compute. Current approaches make approximations to achieve tractability. We propose an approach that expresses information gain in terms of predictive entropies, and apply this method to the Gaussian Process Classifier (GPC). Our approach makes minimal approximations to the full information theoretic objective. Our experimental performance compares favourably to many popular active learning algorithms, and has equal or lower computational complexity. We compare well to decision theoretic approaches also, which are privy to more information and require much more computational time. Secondly, by developing further a reformulation of binary preference learning to a classification problem, we extend our algorithm to Gaussian Process preference learning.

\end{abstract}

\section{Introduction}
In most machine learning systems, the learner passively collects data with which it makes inferences about its environment. In active learning, however, the learner seeks the most useful measurements to be trained upon. The goal of active learning is to produce the best model with the least possible data; this is closely related to the statistical field of optimal experimental design. With the advent of the internet and expansion of storage facilities, vast quantities of unlabelled data have become available, but it can be costly to obtain labels. Finding the most useful data in this vast space calls for efficient active learning algorithms.

Two approaches to active learning are to use decision and information theory \cite{kapoor2007,lindley1956}. The former minimizes the expected losses encountered after making decisions based on the data collected i.e. minimize the Bayes posterior risk \cite{roy2001}. Maximising performance under test is the ultimate objective of most learners, however, evaluating this objective can be very hard. For example, the methods proposed in \cite{kapoor2007,zhu2003} for classification are in general expensive to compute. Furthermore, one may not know the loss function or test distribution in advance, or may want the model to perform well on a variety of loss functions. In extreme scenarios, such as exploratory data analysis, or visualisation, losses may be very hard to quantify. 

This motivates information theoretic approaches to active learning, which are agnostic to the decision task at hand and particular test data, this is known an inductive approach. They seek to reduce the number of feasible models as quickly as possible, using either heuristics (e.g. margin sampling \cite{tong2001}) or by formalising uncertainty using well studied quantities, such as Shannon’s entropy and the KL-divergence \cite{coverandthomas}. Although the latter approach was proposed several decades ago \cite{lindley1956, bernardo1979}, it is not always straightforward to apply the criteria to complicated models such as nonparametric processes with infinite parameter spaces. As a result many algorithms exist which compute approximate posterior entropies, perform sampling, or work with related quantities in non-probabilistic models.

We return to this problem, presenting the full information criterion and demonstrate how to apply it to Gaussian Processes Classification (GPC), yielding a novel active learning algorithm that makes minimal approximations. GPC is a powerful, non-parametric kernel-based model, and poses an interesting problem for information-theoretic active learning because the parameter space is infinite dimensional and the posterior distribution is analytically intractable. We present the information theoretic approach to active learning in Section 2. In Section 3 we apply it to GPC, and show how to extended our method to preference learning. In Section 4 we review other approaches and how they compare to our algorithm. We take particular care to contrast our approach to the Informative Vector Machine, that addresses data point selection for GPs directly. We present results on a wide variety of datasets in Section 5 and conclude in Section 6.

\section{Bayesian Information Theoretic Active Learning}

We consider a fully discriminative model where the goal of active learning is to discover the dependence of some variable $\y\in\mathcal{Y}$ on an input variable $\x\in\mathcal{X}$. The key idea in active learning is that the learner chooses the input queries $\x_i\in\mathcal{X}$ and observes the system's response $\y_i$, rather than passively receiving $(\x_i \y_i)$ pairs.

Within a Bayesian framework we assume existence of some latent parameters, $\param$, that control the dependence between inputs and outputs, $p(\y\vert\x,\param)$. Having observed data $\data = \{(\x_i,\y_i)\}_{i=1}^n$, a posterior distribution over the parameters is inferred, $p(\param|\data)$. The central goal of information theoretic active learning is to reduce the number possible hypotheses maximally fast, i.e. to minimize the uncertainty about the parameters using Shannon's entropy \cite{coverandthomas}. Data points $\mathcal{D}'$ are selected that satisfy $\argmin_{\mathcal{D}'}\rmH[\param|\mathcal{D}']=-\int p(\param|\mathcal{D}')\log p(\param|\mathcal{D}') \mathrm{d}\param$. Solving this problem in general is NP-hard; however, as is common in sequential decision making tasks a myopic (greedy) approximation is made \cite{heckerman1995}. It has been shown that the myopic policy can perform near-optimally \cite{golovin2010,dasgupta2005}. Therefore, the objective is to seek the data point $\x$ that maximises the decrease in expected posterior entropy:

\begin{align}	
	\label{eqn:ent_change}
	\argmax_{\x} \rmH[\param | \data] - \E_{\y\sim p(\y|\x\data)} \left[ \rmH[\param| \y, \x,\data] \right] 
\end{align}

Note that expectation over the unseen output $\y$ is required. Many works e.g. \cite{mackay1992, krishnapuram2004, lawrence2003} propose using this objective directly. However, parameter posteriors are often high dimensional and computing their entropies is usually intractable. Furthermore, for nonparametric processes the parameter space is infinite dimensional so Eqn.\,\eqref{eqn:ent_change} becomes poorly defined. To avoid gridding parameter space (exponentially hard with dimensionality), or sampling (from which it is notoriously hard to estimate entropies without introducing bias \cite{panzeri2007}), these papers make Gaussian or low dimensional approximations and calculate the entropy of the approximate posterior. A second computational difficulty arises; if $N_{\x}$ data points are under consideration, and $N_{\y}$ responses may be seen, then $\mathcal{O}(N_{\x}N_{\y})$, potentially expensive, posterior updates are required to calculate Eqn.\,\eqref{eqn:ent_change}.

An important insight arises if we note that the objective in Eqn.\,\eqref{eqn:ent_change} is equivalent to the conditional mutual information between the unknown output and the parameters, $\rmI[\param,\y\vert\x,\data]$. Using this insight it is simple to show that the objective can be rearranged to compute entropies in $\y$ space:

\begin{align}
\argmax_{\x} \rmH[\y \vert \x, \data] - \E_{\param\sim p(\param|\data)} \left[ \rmH[\y \vert \x,\param] \right] \label{eqn:rearrangement} 
\end{align}

Eqn.\,\eqref{eqn:rearrangement} overcomes the challenges we described for Eqn.\,\eqref{eqn:ent_change}. Entropies are now calculated in, usually low dimensional, output space. For binary classification, these are just entropies of Bernoulli variables. Also $\param$ is now conditioned only on $\data$, so only $\mathcal{O}(1)$ posterior updates are required. Eqn.\,\eqref{eqn:rearrangement} also provides us with an interesting intuition about the objective; we seek the $\x$ for which the model is marginally most uncertain about $\y$ (high $\rmH[\y \vert \x, \data]$), but for which individual settings of the parameters are confident (low $\E_{\param\sim p(\param|\data)} \left[ \rmH[\y \vert \x,\param] \right]$). This can be interpreted as seeking the $\x$ for which the parameters under the posterior disagree about the outcome the most, so we refer to this objective as Bayesian Active Learning by Disagreement (BALD). We present a method to apply Eqn.\,\eqref{eqn:rearrangement} directly to GPC and preference learning. We no longer need to build our entropy calculation around the type of posterior approximation (as in \cite{mackay1992, krishnapuram2004, lawrence2003}) but are free to choose from many of the available algorithms. Minimal additional approximations are introduced, and so, to our knowledge our algorithm represents the most exact and fastest way to perform full information-theoretic active learning in non-parametric discriminative models.

\section{Gaussian Processes for Classification and Preference Learning\label{sec:GPC}}

In this section we derive the BALD algorithm for Gaussian Process classification (GPC). GPs are a powerful and popular non-parametric tool for regression and classification. GPC appears to be an especially challenging problem for information-theoretic active learning because the parameter space is infinite, however, by using \eqref{eqn:rearrangement} we are able to calculate fully the relevant information quantities without having to work out entropies of infinite dimensional objects. The probabilistic model underlying GPC is as follows:
\begin{align}
	&f \sim \mathrm{GP}(\mu(\cdot),k(\cdot,\cdot)) \notag \\
	&\y\vert\x,f \sim\mathrm{Bernoulli}(\Phi(f(\x))) \notag
\end{align}
The latent parameter, now called $f$ is a function $\mathcal{X}\rightarrow\mathbb{R}$, and is assigned a Gaussian process prior with mean $\mu(\cdot)$ and covariance function or kernel $k(\cdot,\cdot)$. We consider the probit case where given the value of $f$, $y$ takes a Bernoulli distribution with probability $\Phi(f(\x))$, and $\Phi$ is the Gaussian CDF. For further details on GPs see \cite{rasmussen2005}.

Inference in the GPC model is intractable; given some observations $\data$, the posterior over $f$ becomes non-Gaussian and complicated. The most commonly used approximate inference methods -- EP,  Laplace approximation, Assumed Density Filtering and sparse methods -- all approximate the posterior by a Gaussian \cite{rasmussen2005}. Throughout this section we will assume that we are provided with such a Gaussian approximation from one of these methods, though the active learning algorithm does not care which one. In our derivation we will use {\scriptsize$\stackrel{1}{\approx}$} to indicate where such an approximation is exploited.

The informativeness of a query $\x$ is computed using Eqn.\,\eqref{eqn:rearrangement}.  The entropy of the binary output variable $\y$ given a fixed $f$ can be expressed in terms of the binary entropy function $\rmh$: 
\begin{align}
\rmH[y\vert\x,f] &= \rmh\left(\Phi(f(\x)\right) \notag\\
\rmh(p)&=- p\log p - (1-p)\log(1-p) \notag
\end{align}
Expectations over the posterior need to be computed. Using a Gaussian approximation to the posterior, for each $\x$, $f_{\x} = f(\x)$ will follow a Gaussian distribution with mean $\mu_{\x,\data}$ and variance $\sigma_{\x,\data}^2$. To compute Eqn.\,\eqref{eqn:rearrangement} we have to compute two entropy quantities. The first term in Eqn.\,\eqref{eqn:rearrangement}, $\rmH[y\vert\x,\data]$ can be handled analytically for the probit case:
\begin{align}
	\rmH[y\vert\x,\data] &\stackrel{1}{\approx} \rmh\left( \int \Phi( f_{\x} )  \mathcal{N}(f_{\x}\vert \mu_{\x,\data},\sigma_{\x,\data}^2) df_{\x} \right) \notag \\ 
	&= \rmh \left( \Phi\left( \frac{\mu_{\x,\data}}{\sqrt{\sigma^2_{\x,\data} + 1}} \right)\right) \label{ent_mean}
\end{align}
The second term, $\E_{f \sim p(f\vert\data)} \left[ \rmH[\y\vert\x, f] \right]$ can be computed approximately as follows:
\begin{align}
	&\E_{f \sim p(f\vert\data)} \left[ \rmH[\y\vert\x, f] \right] \notag\\	 
	&\quad \stackrel{1}{\approx}\int \rmh(\Phi(f_{\x})) \mathcal{N}(f_{\x}\vert \mu_{\x,\data},\sigma_{\x,\data}^2)df_{\x}\label{eqn:mean_entropy}\\
	&\quad\stackrel{2}{\approx} \int \exp\left(-\frac{f_{\x}^2}{\pi\ln2}\right) \mathcal{N}(f_{\x}\vert \mu_{\x,\data},\sigma_{\x,\data}^2)df_{\x}\notag\\	
	&\quad= \frac{C}{\sqrt{\sigma_{\x,\data}^2 + C^2}}\exp\left(-\frac{\mu_{\x,\data}^2}{2\left(\sigma_{\x,\data}^2 + C^2\right)}\right)\notag
\end{align}

where $C=\sqrt{\frac{\pi\ln2}{2}}$. The first approximation, {\scriptsize $\stackrel{1}{\approx}$}, reflects the Gaussian approximation to the posterior. The integral in the left hand side of Eqn.\,\eqref{eqn:mean_entropy} is intractable. By performing a Taylor expansion on $\ln \rmh(\Phi(f_{\x}))$ (see supplementary material) we can see that it can be approximated up to $\mathcal{O}(f_{\x}^4)$ by a squared exponential curve, $\exp(-f_{\x}^2/\pi\ln2)$. We will refer to this approximation as {\scriptsize $\stackrel{2}{\approx}$}. Now we can apply the standard convolution formula for Gaussians to finally get a closed form expression for both terms of Eqn.\,\eqref{eqn:rearrangement}.

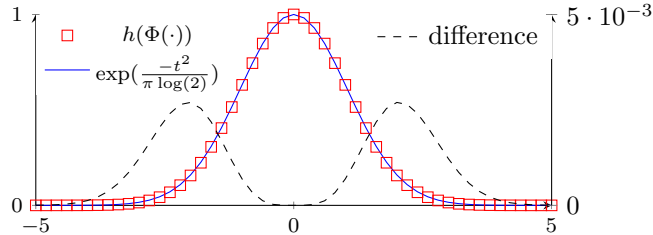
\begin{figure}\centering
%
%
\begin{tikzpicture}

\begin{axis}[%
footnotesize,
scale only axis,
width=2.7in,
height=1.0in,
xmin=-5, xmax=5,
ymin=0, ymax=1,
xtick={-5,0,5},
ytick = {0,1},
axis y line = left,
axis x line = bottom,
legend style={ at={(0,1)}, anchor=north west, draw = none}]
]

\addplot [
color=red,
only marks,
mark=square,
mark options={solid}
]
coordinates{ (-5,6.64369e-06) (-4.8,1.72218e-05) (-4.6,4.2873e-05) (-4.4,0.000102503) (-4.2,0.000235365) (-4,0.000519064) (-3.8,0.0010995) (-3.6,0.00223711) (-3.4,0.00437257) (-3.2,0.00821083) (-3,0.0148147) (-2.8,0.0256873) (-2.6,0.0428103) (-2.4,0.0685917) (-2.2,0.105681) (-2,0.156615) (-1.8,0.223311) (-1.6,0.306444) (-1.4,0.40484) (-1.2,0.515021) (-1,0.631083) (-0.8,0.745014) (-0.6,0.847502) (-0.4,0.929133) (-0.2,0.981797) (0,1) (0.2,0.981797) (0.4,0.929133) (0.6,0.847502) (0.8,0.745014) (1,0.631083) (1.2,0.515021) (1.4,0.40484) (1.6,0.306444) (1.8,0.223311) (2,0.156615) (2.2,0.105681) (2.4,0.0685917) (2.6,0.0428103) (2.8,0.0256873) (3,0.0148147) (3.2,0.00821083) (3.4,0.00437257) (3.6,0.00223711) (3.8,0.0010995) (4,0.000519064) (4.2,0.000235365) (4.4,0.000102503) (4.6,4.2873e-05) (4.8,1.72218e-05) (5,6.64369e-06)
};
\label{plots:approx_true}
\addlegendentry{$h(\Phi(\cdot))$}

\addplot [
color=blue,
solid
]
coordinates{ (-5,1.03285e-05) (-4.8,2.54061e-05) (-4.6,6.02395e-05) (-4.4,0.00013768) (-4.2,0.000303323) (-4,0.000644146) (-3.8,0.00131859) (-3.6,0.00260182) (-3.4,0.0049487) (-3.2,0.00907298) (-3,0.0160344) (-2.8,0.0273151) (-2.6,0.0448534) (-2.4,0.0709961) (-2.2,0.108322) (-2,0.159311) (-1.8,0.22585) (-1.6,0.30863) (-1.4,0.406537) (-1.2,0.516189) (-1,0.631774) (-0.8,0.745348) (-0.6,0.847622) (-0.4,0.929159) (-0.2,0.981799) (0,1) (0.2,0.981799) (0.4,0.929159) (0.6,0.847622) (0.8,0.745348) (1,0.631774) (1.2,0.516189) (1.4,0.406537) (1.6,0.30863) (1.8,0.22585) (2,0.159311) (2.2,0.108322) (2.4,0.0709961) (2.6,0.0448534) (2.8,0.0273151) (3,0.0160344) (3.2,0.00907298) (3.4,0.0049487) (3.6,0.00260182) (3.8,0.00131859) (4,0.000644146) (4.2,0.000303323) (4.4,0.00013768) (4.6,6.02395e-05) (4.8,2.54061e-05) (5,1.03285e-05)
};
\label{plots:approx_approx}
\addlegendentry{$\exp(\frac{-t^2}{\pi\log(2)})$}

\end{axis}

\begin{axis}[%
scale only axis,
width=2.7in,
height=1.0in,
xmin=-5, xmax=5,
ymin=0, ymax=0.005,
xtick={-5,0,5},
ytick = {0,0.005},
axis y line = right,
axis x line = none,
legend style={ at={(1,1)}, anchor=north east, draw = none}]
]

\addplot [
color=black,
dashed
]
coordinates{ (-5,3.68482e-06) (-4.8,8.18427e-06) (-4.6,1.73665e-05) (-4.4,3.51775e-05) (-4.2,6.79579e-05) (-4,0.000125081) (-3.8,0.000219088) (-3.6,0.000364708) (-3.4,0.000576126) (-3.2,0.000862153) (-3,0.00121977) (-2.8,0.00162773) (-2.6,0.00204316) (-2.4,0.00240434) (-2.2,0.0026417) (-2,0.00269602) (-1.8,0.00253851) (-1.6,0.00218519) (-1.4,0.00169772) (-1.2,0.00116807) (-1,0.000690885) (-0.8,0.000334143) (-0.6,0.000120243) (-0.4,2.60299e-05) (-0.2,1.71855e-06) (0,-0) (0.2,1.71855e-06) (0.4,2.60299e-05) (0.6,0.000120243) (0.8,0.000334143) (1,0.000690885) (1.2,0.00116807) (1.4,0.00169772) (1.6,0.00218519) (1.8,0.00253851) (2,0.00269602) (2.2,0.0026417) (2.4,0.00240434) (2.6,0.00204316) (2.8,0.00162773) (3,0.00121977) (3.2,0.000862153) (3.4,0.000576126) (3.6,0.000364708) (3.8,0.000219088) (4,0.000125081) (4.2,6.79579e-05) (4.4,3.51775e-05) (4.6,1.73665e-05) (4.8,8.18427e-06) (5,3.68482e-06)
};
\label{plots:approx_error}
\addlegendentry{difference}

\end{axis}
\end{tikzpicture}
\caption{Analytic approximation ({\scriptsize $\stackrel{1}{\approx}$}) to the binary entropy of the error function (\ref{plots:approx_true}) by a squared exponential (\ref{plots:approx_approx}). The absolute error (\ref{plots:approx_error}) remains under $3\cdot 10^{-3}$. }\label{fig:trick}
\end{figure}

Fig.\,\ref{fig:trick} depicts the striking accuracy of this simple approximation. The maximum possible error that will be incurred when using this approximation is if $\mathcal{N}(f_{\x}\vert \mu_{\x,\data},\sigma_{\x,\data}^2)$ is centred at $\mu_{\x,\data}=\pm 2.05$  with $\sigma_{\x,\data}^2$ tending to zero (see Fig.\,\ref{fig:trick}, absolute error \ref{plots:approx_error}), yielding only a 0.27\% error in the integral in Eqn.\,\eqref{eqn:mean_entropy}. The authors are unaware of previous use of this simple and useful approximation in this context.  In Section \ref{sec:experiments} we investigate experimentally the information lost from approximations {\scriptsize $\stackrel{1}{\approx}$} and {\scriptsize $\stackrel{2}{\approx}$} as compared to the golden standard of extensive Monte Carlo simulation.

To summarise, the BALD algorithm for Gaussian process classification consists of two steps. First it applies any standard approximate inference algorithm for GPCs (such as EP) to obtain the posterior predictive mean $\mu_{\x,\data}$ and $\sigma_{\x,\data}$ for each point of interest $\x$. Then, it selects a query $\x$ that maximises the following objective function:

\begin{equation}
	\rmh \left( \Phi\left( \frac{\mu_{\x,\data}}{\sqrt{\sigma^2_{\x,\data} + 1}} \right)\right) - \frac{C \exp\left(-\frac{\mu_{\x,\data}^2}{2\left(\sigma_{\x,\data}^2 + C^2\right)}\right)}{\sqrt{\sigma_{\x,\data}^2 + C^2}} \label{eqn:BALD_GPC}
\end{equation}

For most practically relevant kernels, the objective \eqref{eqn:BALD_GPC} is a smooth and differentiable function of $\x$, so gradient-based optimisation procedures can be used to find the maximally informative query.

\subsection{Extension: Learning Hyperparameters \label{sec:hyperparameters}}

In many applications the parameter set $\param$ naturally divides into parameters of interest, $\param^+$, and nuisance parameters $\param^-$, i.e. $\param=\{\param^+,\param^-\}$. In such settings, the active learning may want to query points that are maximally informative about $\param^+$, while not caring about $\param^-$. By integrating Eqn.\,\eqref{eqn:ent_change} over the nuisance parameters, $\param^-$, BALD's objective is re-derived as:

\begin{align} 
&\rmH\left[\E_{p(\param^+,\param^-\vert \data)}\left[\y|\x,\param^+,\param^-\right]\right] \notag\\
&\quad- \E_{p(\param^+|\data)} \left[ \rmH\left[\E_{p(\param^-|\param^+,\data)}[ \y \vert \x, \param^+,\param^- ]\right] \right]\label{eqn:BALD_bipartite}
\end{align}

In the context of GP models, hyperparameters typically control the smoothness or spatial length-scale of functions. If we maintain a posterior distribution over these hyperparameters, which we can do e.\,g.\ via Hamiltonian Monte Carlo, we can choose either to treat them as nuisance parameters $\param^-$ and use Eq.\ \ref{eqn:BALD_bipartite}, or to include them in $\param^+$ and perform active learning over them as well. In certain cases, such as automatic relevance determination \cite{rasmussen2005}, it may even make sense to treat hyperparameters as variables of primary interest, and the function $f$ itself as nuisance parameter $\param^-$.

\subsection{Preference Learning}

Our active learning framework for GPC can be extended to the important problem of preference learning \cite{furnkranz2003, chu2005}. In preference learning the dataset consists for pairs of items $(\upref_i,\vpref_i)\in\mathcal{X}^2$ with binary labels, $y_i\in\{0,1\}$. $y_i=1$ means instance $\upref_i$ is preferred to $\vpref_i$, denoted $\upref_i\succ \vpref_i$. The task is to predict the preference relation between any $(\upref,\vpref)$. We can view this as a special case of building a classifier on pairs of inputs $\rmh:\mathcal{X}^2\mapsto\{0,1\}$. \cite{chu2005} propose a Bayesian approach, using a latent preference function $f$, over which a GP prior is defined. The model predicts preference,  $\upref_i \succ \vpref_i$ whenever $f(\upref_i)+\epsilon_{u_i}>f(\vpref_i)+\epsilon_{v_i}$, where $\epsilon_{u_i}, \epsilon_{v_i}$ denote additive Gaussian noise. Under this model, the likelihood of $f$ becomes:

\begin{align}
	\mathbb{P}[y=1\vert (\upref_i,\vpref_i), f] &= \mathbb{P}[\upref_i\succ \vpref_i \vert f] \notag\\
	&=  \Phi\left(\frac{f(\upref_i) - f(\vpref_i)}{\sqrt{2}\sigma_{noise}}\right)
\end{align}

By rescaling the latent function $f$, it can be assumed w.l.o.g.\,that $\sqrt{2}\sigma_{noise}=1$. The likelihood only depends on the difference between $f(\upref)$ and $f(\vpref)$. We therefore define $g(\upref,\vpref)=f(\upref)-f(\vpref)$, and do inference entirely in terms of $g$, for which the likelihood becomes the same as for probit classification: $y|\upref,\vpref,f\sim \mathrm{Bernoulli}(\Phi(g(\upref,\vpref)))$. We observe that a GP prior is induced on $g$ because it is formed by performing a linear operation on $f$, for which we have a GP prior already $f\sim \mathrm{GP}(0,k)$. We can derive the induced covariance function of $g$ as (derivation in the Supplementary material) as: $k_{\mathrm{pref}}((\upref_i,\vpref_i),(\upref_j,\vpref_j)) = k(\upref_i,\upref_j) + k(\vpref_i,\vpref_j) - k(\upref_i,\vpref_j) - k(\vpref_i,\upref_j)$.

Note that this kernel $k_{\mathrm{pref}}$ respects the anti-symmetry properties desired for a preference learning scenario, i.e. the value $g(u,v)$ is perfectly anti-correlated with $g(v,u)$, ensuring $\mathbb{P}[\upref\succ \vpref] = 1 - \mathbb{P}[\vpref \succ \upref]$ holds. Thus, we can conclude that the GP preference learning framework of \cite{chu2005}, is equivalent to GPC with a particular class of kernels, that we may call the \emph{preference judgement kernels}. Therefore, our active learning algorithm presented in Section \ref{sec:GPC} for GPC can readily be applied to pairwise preference learning also.

\section{Related Methodologies}

There are a number of closely related algorithms for active classification which we now review.

\paragraph{The Informative Vector Machine (IVM):} Perhaps the most closely related approach is the IVM \cite{lawrence2003}. This popular,and successful approach to active learning was designed specifically for GPs; it uses an information theoretic approach and so appears very similar to BALD. The IVM algorithm was designed for subsampling a dataset for training a GP, so it is privy to the $\y$ values before including a measurement; it cannot therefore work explicitly in output space i.e. with Eqn.\,\eqref{eqn:rearrangement}. The IVM uses Eqn.\,\eqref{eqn:ent_change}, but parameter entropies are calculated approximately in the marginal subspace corresponding to the observed data points. The entropy decrease after inclusion of a new data point can then be calculated efficiently using the GP covariance matrix.

Although the IVM and BALD are motivated by the same objective, they work fundamentally differently when approximate inference is carried out. At any time both methods have an approximate posterior $q_t(\param|\data)$, this can be updated with the likelihood of a new data point $p(y_{t+1}|f,\x_{t+1})$, yielding $\hat{p}_{t+1}(\param|\data,\x_{t+1}, y_{t+1})=\frac{1}{Z}q_t(\param|\data)p(y_{t+1}|f,\x_{t+1})$. If the posterior at $t+1$ is approximated directly one gets $q_{t+1}(\param|\data,\x_{t+1},\y_{t+1})$. BALD calculates the entropy difference between $q_t$ and $\hat{p}_{t+1}$, without having to compute $q_{t+1}$ for each candidate $\x$. In contrast, the IVM calculates the entropy change between $q_{t}$ and $q_{t+1}$. The IVM's approach cannot calculate the entropy of the full infinite dimensional posterior, and requires $\mathcal{O}(N_{\x}N_{\y})$ posterior updates. To do these updates efficiently, approximate inference is performed using Assumed Density Filtering (ADF). Using ADF means that $q_{t+1}$ is a direct approximation to $\hat{p}_{t+1}$, indicating that the IVM makes a further approximation to BALD. Since BALD only requires $\mathcal{O}(1)$ posterior updates it can afford to use more accurate, iterative procedures, such as EP.

\paragraph{Information Theoretic approaches:} Maximum Entropy Sampling (MES) \cite{sebastiani2000} explicitly works in dataspace (Eqn.\,\eqref{eqn:rearrangement}). MES was proposed for regression models with input-independent observation noise. Although Eqn.\,\eqref{eqn:rearrangement} is used, the second term is constant because of input independent noise and is ignored. One cannot, however, use MES for heteroscedastic regression or classification; it fails to differentiate between model uncertainty and observation uncertainty (about which our model may be confident). Some toy demonstrations show this `information based' active learning criterion performing pathologically in classification by repeatedly querying points close the decision boundary or in regions of high observation uncertainty  e.g. \cite{huang2010}. This is because MES is inappropriate in this domain; BALD distinguishes between observation and model uncertainty and eliminates these problems as we will show.

Mutual-information based objective functions are presented in \cite{ertin2003,fuhrmann2003}. They maximise the mutual information between the variable being measured and the variable of interest. Fuhrmann \cite{fuhrmann2003} applies this to linear Gaussian models and acoustic arrays, Ertin \emph{et al.} \cite{ertin2003} to a communications channel. Although related, these objectives do not work with the model parameters and are not applied to classification. \cite{guestrin2005, krause2006} also use mutual information. They specify interest points in advance and maximise the expected mutual information between the predictive distributions at these points and at the observed locations. Although this is a objective is promising for regression, it is not tractable for models with input-dependent observation noise, such as classification or preference learning.


\paragraph{Decision theoretic:} We briefly mention decision theoretic approaches to active learning. Two closely related algorithms, \cite{kapoor2007, zhu2003}, seek to minimize the expected cost i.e. loss weighted misclassification probability on all seen and future data. These methods observe the locations of the test points and their objective functions become monotonic in the predictive entropies at the test points. \cite{kapoor2007} also includes an empirical error term that can yield pathological behaviour (we investigate this experimentally). These approaches are computationally expensive, requiring $\mathcal{O}(N_{\x}N_{\y})$ posterior updates. Also, they must know the locations of the test data (and thus are transductive approaches); designing an inductive, decision-theoretic algorithm  is an open, hard problem as it would require expensive integration over possible test data distributions.

\paragraph{Non-probabilistic} Some non-probabilistic methods have close analogues to information theoretic active learning. Perhaps the most ubiquitous is active learning for SVMs \cite{tong2001,seung1992}, where the volume of Version Space (VS) is used as a proxy for the posterior entropy. If a uniform (improper) prior is used with a deterministic classification likelihood, the log volume of VS and Bayesian posterior entropy are in fact equivalent. Just as Bayesian posteriors become intractable after observing many data points, VS can become complicated. \cite{tong2001} proposes methods for approximating VS with a simple shapes, such as hyperspheres (their simplest approximation reduces to margin sampling). This closely resembles approximating a Bayesian posterior using a Gaussian distribution via the Laplace or EP approximations. \cite{seung1992} sidesteps the problem by working with predictions. The algorithm, Query by Committee (QBC), samples parameters from VS (committee members), they vote on the outcome of each possible $\x$. The $\x$ with the most balanced vote is selected; this is termed the `principle of maximal disagreement'. If BALD is used with a sampled posterior, query by committee is implemented but with a probabilistic measure of disagreement. QBC's deterministic vote criterion discards confidence in the predictions and so can exhibit the same pathologies as MES.


\begin{figure}[t]\centering
\begin{tabular}{|c|c|c|c|}
\hline
&MCMC&EP ($\stackrel{1}{\approx}$)&Laplace ($\stackrel{1}{\approx}$)\\ \hline
\hline
MC & 0 & $7.51\pm2.51$ & $41.57\pm4.02$ \\
$\stackrel{2}{\approx}$ & $0.16\pm0.05$ & $7.43\pm2.40$ & $40.45\pm3.67$ \\ \hline
\end{tabular}
\caption{Percentage approximation error ($\pm$1 s.d.) for different methods of approximate inference (\emph{columns}) and approximation methods for evaluating Eqn.\,\eqref{eqn:mean_entropy} (\emph{rows}). The results indicate that {\scriptsize $\stackrel{2}{\approx}$} is a very accurate approximation; EP causes some loss and Laplace significantly more, which is in line with the comparison presented in \cite{Kuss05}. For our experiments we use EP.}\label{fig:trick_results}
\end{figure}

\section{Experiments} \label{sec:experiments}

\begin{figure*}[t]
\noindent\makebox[\textwidth]{%
\begin{tabular}{ccc}
%
%
\begin{tikzpicture}

\begin{axis}[
footnotesize,
width= 1.6in,
height= 1.6in,
xmin=0, xmax=30,
ymin=0, ymax=30,
title={(a) block in the middle},
ytick={0,7.5,15,22.5,30},
xtick = {0,7.5,15,22.5,30},
xlabel = {Dim. 1},
ylabel = {Dim. 2},
xticklabels={,,,,},
yticklabels={,,,,},
axis on top,
axis y line = left,
axis x line = bottom
]
\addplot [
color=red,
only marks,
mark=*,
mark options={scale = 0.4}
]
coordinates{ (13,13) (15,13) (15.5,13) (16.5,13) (13.5,13.5) (15,13.5) (16,13.5) (16.5,13.5) (13,14) (13.5,14) (14,14) (15.5,14) (17,14) (13,14.5) (14,14.5) (15.5,14.5) (16,14.5) (16.5,14.5) (13,15) (15,15) (16,15) (16.5,15) (17,15) (13,15.5) (13.5,15.5) (14,15.5) (15.5,15.5) (13,16) (13.5,16) (14.5,16) (15,16) (16.5,16) (13,16.5) (14.5,16.5) (15.5,16.5) (16,16.5) (13,17) (13.5,17) (14,17) (14.5,17) (15,17) (15.5,17) (16,17) (8.86203,21.9844) (1.87844,22.0987) (7.62888,6.41534) (5.50559,21.3585) (7.97933,22.783) (6.66413,14.0934) (2.66314,8.77752) (14.6582,0.0330902) (13.6696,8.46285) (2.74037,5.67331) (13.6087,28.3003) (12.3521,14.9273) (3.36577,27.7504) (14.5511,5.06006)
};
\label{plots:negatives}

\addplot [
color=black,
only marks,
mark=square*,
mark options={scale = 0.4}
]
coordinates{ (13.5,13) (14,13) (14.5,13) (16,13) (17,13) (13,13.5) (14,13.5) (14.5,13.5) (15.5,13.5) (17,13.5) (14.5,14) (15,14) (16,14) (16.5,14) (13.5,14.5) (14.5,14.5) (15,14.5) (17,14.5) (13.5,15) (14,15) (14.5,15) (15.5,15) (14.5,15.5) (15,15.5) (16,15.5) (16.5,15.5) (17,15.5) (14,16) (15.5,16) (16,16) (17,16) (13.5,16.5) (14,16.5) (15,16.5) (16.5,16.5) (17,16.5) (16.5,17) (17,17) (18.3773,17.3043) (24.9794,19.4896) (26.7589,22.3496) (17.8888,10.3495) (16.4577,2.38458) (20.4057,11.7607) (19.0268,2.51172) (20.4262,16.5161) (28.71,23.0737) (21.1403,3.08034) (16.3379,4.55224) (20.1516,20.3166) (23.7032,25.0623) (26.6266,28.273) (16.6152,29.476) (28.6434,7.37318)
};
\label{plots:positives}

\end{axis}
\end{tikzpicture}&
%
%
\begin{tikzpicture}

\begin{axis}[
footnotesize,
width= 1.6in,
height= 1.6in,
xmin=0, xmax=30,
ymin=0, ymax=30,
title={(b) block in the corner},
xlabel = {Dim. 1},
ylabel = {Dim. 2},
ytick={0,7.5,15,22.5,30},
xtick = {0,7.5,15,22.5,30},
xticklabels={,,,,},
yticklabels={,,,,},
axis on top,
axis y line = left,
axis x line = bottom
]

\addplot [
color=red,
only marks,
mark=*,
mark options={scale=0.4}
]
coordinates{ (1,1) (1.5,1) (2,1) (2.5,1) (3,1) (3.5,1) (4,1) (4.5,1) (5,1) (1,1.5) (1.5,1.5) (2,1.5) (2.5,1.5) (3,1.5) (3.5,1.5) (4,1.5) (4.5,1.5) (5,1.5) (1,2) (1.5,2) (2,2) (2.5,2) (3,2) (3.5,2) (4,2) (4.5,2) (5,2) (1,2.5) (1.5,2.5) (2,2.5) (2.5,2.5) (3,2.5) (3.5,2.5) (4,2.5) (4.5,2.5) (5,2.5) (1,3) (1.5,3) (2,3) (2.5,3) (3,3) (3.5,3) (4,3) (4.5,3) (5,3) (1,3.5) (1.5,3.5) (2,3.5) (2.5,3.5) (3,3.5) (3.5,3.5) (4,3.5) (4.5,3.5) (5,3.5) (1,4) (1.5,4) (2,4) (2.5,4) (3,4) (3.5,4) (4,4) (4.5,4) (5,4) (1,4.5) (1.5,4.5) (2,4.5) (2.5,4.5) (3,4.5) (3.5,4.5) (4,4.5) (4.5,4.5) (5,4.5) (1,5) (1.5,5) (2,5) (2.5,5) (3,5) (3.5,5) (4,5) (4.5,5) (5,5) (14.7594,22.137) (9.50211,21.9321) (11.9146,15.4247) (4.42475,17.1991) (7.09258,9.45929) (11.0419,22.1845) (14.5021,12.2209) (12.9235,13.3655) (10.8308,29.5065) (10.2906,19.3776) (4.33421,3.57372) (10.1417,15.6132) (6.28981,20.1691) (8.16591,16.8388)
};

\addplot [
color=black,
only marks,
mark=square*,
mark options={scale=0.4}
]
coordinates{ (17.1549,13.4394) (29.7725,16.7571) (16.7772,22.7654) (28.2901,17.0776) (16.8774,8.28907) (17.5203,9.19147) (27.1422,7.43214) (16.8312,5.41786) (20.3286,9.78169) (24.4829,8.00663) (27.1345,0.366002) (29.6109,25.1893) (24.263,16.1932) (15.2474,14.4045) (26.6495,29.111) (16.7822,3.87225)
};

\end{axis}
\end{tikzpicture}&
%
%
\begin{tikzpicture}

\begin{axis}[
footnotesize,
width= 1.6in,
height= 1.6in,
xmin=0, xmax=30,
ymin=0, ymax=30,
title={(c) checkerboard},
xlabel = {Dim. 1},
ylabel = {Dim. 2},
ytick={0,7.5,15,22.5,30},
xtick = {0,7.5,15,22.5,30},
xticklabels={,,,,},
yticklabels={,,,,},
axis on top,
axis y line = left,
axis x line = bottom
]

\addplot [
color=red,
only marks,
mark=*,
mark options={scale=0.4}
]
coordinates{ (1,1) (2,1) (3,1) (4,1) (17,1) (18,1) (19,1) (20,1) (1,2) (2,2) (3,2) (4,2) (17,2) (18,2) (19,2) (20,2) (1,3) (2,3) (3,3) (4,3) (17,3) (18,3) (19,3) (20,3) (1,4) (2,4) (3,4) (4,4) (17,4) (18,4) (19,4) (20,4) (9,9) (10,9) (11,9) (12,9) (25,9) (26,9) (27,9) (28,9) (9,10) (10,10) (11,10) (12,10) (25,10) (26,10) (27,10) (28,10) (9,11) (10,11) (11,11) (12,11) (25,11) (26,11) (27,11) (28,11) (9,12) (10,12) (11,12) (12,12) (25,12) (26,12) (27,12) (28,12) (1,17) (2,17) (3,17) (4,17) (17,17) (18,17) (19,17) (20,17) (1,18) (2,18) (3,18) (4,18) (17,18) (18,18) (19,18) (20,18) (1,19) (2,19) (3,19) (4,19) (17,19) (18,19) (19,19) (20,19) (1,20) (2,20) (3,20) (4,20) (17,20) (18,20) (19,20) (20,20) (9,25) (10,25) (11,25) (12,25) (25,25) (26,25) (27,25) (28,25) (9,26) (10,26) (11,26) (12,26) (25,26) (26,26) (27,26) (28,26) (9,27) (10,27) (11,27) (12,27) (25,27) (26,27) (27,27) (28,27) (9,28) (10,28) (11,28) (12,28) (25,28) (26,28) (27,28) (28,28)
};

\addplot [
color=black,
only marks,
mark=square*,
mark options={scale=0.4}
]
coordinates{ (9,1) (10,1) (11,1) (12,1) (25,1) (26,1) (27,1) (28,1) (9,2) (10,2) (11,2) (12,2) (25,2) (26,2) (27,2) (28,2) (9,3) (10,3) (11,3) (12,3) (25,3) (26,3) (27,3) (28,3) (9,4) (10,4) (11,4) (12,4) (25,4) (26,4) (27,4) (28,4) (1,9) (2,9) (3,9) (4,9) (17,9) (18,9) (19,9) (20,9) (1,10) (2,10) (3,10) (4,10) (17,10) (18,10) (19,10) (20,10) (1,11) (2,11) (3,11) (4,11) (17,11) (18,11) (19,11) (20,11) (1,12) (2,12) (3,12) (4,12) (17,12) (18,12) (19,12) (20,12) (9,17) (10,17) (11,17) (12,17) (25,17) (26,17) (27,17) (28,17) (9,18) (10,18) (11,18) (12,18) (25,18) (26,18) (27,18) (28,18) (9,19) (10,19) (11,19) (12,19) (25,19) (26,19) (27,19) (28,19) (9,20) (10,20) (11,20) (12,20) (25,20) (26,20) (27,20) (28,20) (1,25) (2,25) (3,25) (4,25) (17,25) (18,25) (19,25) (20,25) (1,26) (2,26) (3,26) (4,26) (17,26) (18,26) (19,26) (20,26) (1,27) (2,27) (3,27) (4,27) (17,27) (18,27) (19,27) (20,27) (1,28) (2,28) (3,28) (4,28) (17,28) (18,28) (19,28) (20,28)
};

\end{axis}
\end{tikzpicture}\\
%
%
\begin{tikzpicture}

\definecolor{mycolor1}{rgb}{0,0.4,0}
\definecolor{mycolor2}{rgb}{0,1,1}
\definecolor{mycolor3}{rgb}{1,0,1}
\definecolor{mycolor4}{rgb}{1,0.8,0.5}
\definecolor{mycolor5}{rgb}{0.7,0.4,0.01}

\begin{axis}[
footnotesize,
width= 1.6in,
height= 1.6in,
xmin=0, xmax=40,
ymin=0.1, ymax=0.9,
ytick={0.5,0.9},
xtick = {0,20,40},
xlabel = {No. queried points},
ylabel = {Accuracy},
axis on top,
axis y line = left,
axis x line = bottom
]
\addplot [
color=black,
solid,
line width=2.0pt
]
coordinates{ (1,0.450769) (2,0.627845) (3,0.647265) (4,0.712941) (5,0.741003) (6,0.762954) (7,0.773018) (8,0.789136) (9,0.798441) (10,0.806265) (11,0.817283) (12,0.82484) (13,0.827751) (14,0.825769) (15,0.832678) (16,0.83974) (17,0.837371) (18,0.837804) (19,0.838604) (20,0.842851) (21,0.844002) (22,0.84399) (23,0.846894) (24,0.844338) (25,0.84367) (26,0.845705) (27,0.845233) (28,0.844348) (29,0.844579) (30,0.844709) (31,0.843005) (32,0.844029) (33,0.843528) (34,0.843761) (35,0.84444) (36,0.843425) (37,0.842839) (38,0.844874) (39,0.842366) (40,0.844488)
};
\label{plots:BALD}

\addplot [
color=red,
dashed,
line width=1.0pt
]
coordinates{ (1,0.451086) (2,0.49816) (3,0.549225) (4,0.592272) (5,0.599452) (6,0.629475) (7,0.644924) (8,0.640655) (9,0.66117) (10,0.663051) (11,0.670583) (12,0.681396) (13,0.711317) (14,0.708778) (15,0.721724) (16,0.72956) (17,0.733506) (18,0.736772) (19,0.745888) (20,0.747867) (21,0.74745) (22,0.753887) (23,0.754945) (24,0.758861) (25,0.762162) (26,0.763886) (27,0.764308) (28,0.768394) (29,0.773301) (30,0.775483) (31,0.784175) (32,0.781753) (33,0.781804) (34,0.784162) (35,0.784017) (36,0.78191) (37,0.782467) (38,0.787036) (39,0.784253) (40,0.781836)
};
\label{plots:rand}

\addplot [
color=green,
dashed,
line width=1.0pt
]
coordinates{ (1,0.423574) (2,0.507584) (3,0.493758) (4,0.508564) (5,0.498024) (6,0.511173) (7,0.50287) (8,0.509933) (9,0.514595) (10,0.513477) (11,0.517124) (12,0.514896) (13,0.51485) (14,0.520506) (15,0.510279) (16,0.517637) (17,0.512111) (18,0.522211) (19,0.518124) (20,0.512904) (21,0.509206) (22,0.518069) (23,0.505895) (24,0.515942) (25,0.500831) (26,0.514965) (27,0.496354) (28,0.5073) (29,0.496602) (30,0.504057) (31,0.499562) (32,0.500832) (33,0.498164) (34,0.489265) (35,0.504054) (36,0.491417) (37,0.496447) (38,0.492809) (39,0.495733) (40,0.494792)
};
\label{plots:IVM}

\addplot [
color=mycolor1,
dashed,
line width=1.0pt
]
coordinates{ (1,0.423859) (2,0.560142) (3,0.663338) (4,0.691968) (5,0.697218) (6,0.718702) (7,0.724501) (8,0.744462) (9,0.743842) (10,0.745216) (11,0.758551) (12,0.756943) (13,0.759429) (14,0.768432) (15,0.771157) (16,0.771218) (17,0.773911) (18,0.776606) (19,0.775208) (20,0.775321) (21,0.774416) (22,0.776446) (23,0.773276) (24,0.773258) (25,0.775169) (26,0.772137) (27,0.774075) (28,0.773772) (29,0.775545) (30,0.780022) (31,0.780774) (32,0.780143) (33,0.780392) (34,0.77998) (35,0.780831) (36,0.781354) (37,0.778975) (38,0.777965) (39,0.778391) (40,0.777663)
};
\label{plots:maxent}

\addplot [
color=mycolor2,
dashed,
line width=1.0pt
]
coordinates{ (1,0.437744) (2,0.457254) (3,0.557851) (4,0.607051) (5,0.65399) (6,0.683368) (7,0.689059) (8,0.680439) (9,0.685365) (10,0.699959) (11,0.702562) (12,0.717143) (13,0.709763) (14,0.718033) (15,0.717221) (16,0.72375) (17,0.721815) (18,0.727367) (19,0.727636) (20,0.733762) (21,0.740265) (22,0.747369) (23,0.749206) (24,0.746662) (25,0.744187) (26,0.743369) (27,0.743917) (28,0.745755) (29,0.751835) (30,0.75345) (31,0.754406) (32,0.753725) (33,0.763583) (34,0.768715) (35,0.770087) (36,0.771398) (37,0.7718) (38,0.775158) (39,0.778767) (40,0.78049)
};
\label{plots:QBC2}

\addplot [
color=blue,
dashed,
line width=1.0pt
]
coordinates{ (1,0.417143) (2,0.559673) (3,0.652982) (4,0.679753) (5,0.693065) (6,0.709019) (7,0.710244) (8,0.723463) (9,0.731474) (10,0.741036) (11,0.734572) (12,0.740239) (13,0.741968) (14,0.740834) (15,0.736291) (16,0.746526) (17,0.752956) (18,0.751852) (19,0.75495) (20,0.754615) (21,0.756663) (22,0.762286) (23,0.762192) (24,0.763659) (25,0.759825) (26,0.75857) (27,0.76031) (28,0.763636) (29,0.767371) (30,0.762333) (31,0.759305) (32,0.758561) (33,0.759312) (34,0.760122) (35,0.758054) (36,0.756994) (37,0.755904) (38,0.761444) (39,0.761644) (40,0.765411)
};
\label{plots:QBC100}

\addplot [
color=mycolor3,
dashed,
line width=1.0pt
]
coordinates{ (1,0.5386) (2,0.565211) (3,0.634337) (4,0.700194) (5,0.725775) (6,0.741928) (7,0.752486) (8,0.760961) (9,0.77014) (10,0.767481) (11,0.772995) (12,0.78203) (13,0.788191) (14,0.790413) (15,0.792089) (16,0.794954) (17,0.799203) (18,0.801811) (19,0.800656) (20,0.805692) (21,0.801845) (22,0.803726) (23,0.808346) (24,0.807935) (25,0.812626) (26,0.815763) (27,0.815573) (28,0.822299) (29,0.823795) (30,0.826076) (31,0.825052) (32,0.827026) (33,0.828653) (34,0.828648) (35,0.827402) (36,0.829904) (37,0.82963) (38,0.830456) (39,0.833531) (40,0.836065)
};
\label{plots:SVM}

\addplot [
color=mycolor4,
dashed,
line width=1.0pt
]
coordinates{ (1,0.432646) (2,0.4132) (3,0.40679) (4,0.463525) (5,0.488598) (6,0.462491) (7,0.472777) (8,0.45572) (9,0.45679) (10,0.447679) (11,0.438245) (12,0.440148) (13,0.434083) (14,0.430394) (15,0.424877) (16,0.421133) (17,0.417703) (18,0.414181) (19,0.41102) (20,0.409248) (21,0.407317) (22,0.405129) (23,0.403463) (24,0.401783) (25,0.400397) (26,0.399241) (27,0.397389) (28,0.39576) (29,0.394194) (30,0.392164) (31,0.391381) (32,0.389456) (33,0.387728) (34,0.386783) (35,0.385202) (36,0.383672) (37,0.382133) (38,0.38079) (39,0.378258) (40,0.376427)
};
\label{plots:dec}

\addplot [
color=mycolor5,
dashed,
line width=1.0pt
]
coordinates{ (1,0.423336) (2,0.491499) (3,0.583662) (4,0.625896) (5,0.661889) (6,0.667475) (7,0.683743) (8,0.694363) (9,0.711161) (10,0.711068) (11,0.712454) (12,0.728513) (13,0.729423) (14,0.730234) (15,0.743882) (16,0.748153) (17,0.750597) (18,0.758435) (19,0.760625) (20,0.768514) (21,0.768946) (22,0.767941) (23,0.772845) (24,0.779496) (25,0.772675) (26,0.77328) (27,0.775972) (28,0.773226) (29,0.775773) (30,0.78041) (31,0.779872) (32,0.780489) (33,0.779693) (34,0.780463) (35,0.784821) (36,0.787482) (37,0.784063) (38,0.785672) (39,0.785049) (40,0.791106)
};
\label{plots:semi}

\addplot [
color=gray,
dashed,
line width=1.0pt
]
coordinates{ (1,0.429559) (2,0.382814) (3,0.3505) (4,0.328911) (5,0.313894) (6,0.298766) (7,0.287718) (8,0.277128) (9,0.268509) (10,0.261305) (11,0.254163) (12,0.248497) (13,0.243079) (14,0.238635) (15,0.233852) (16,0.229898) (17,0.225825) (18,0.222464) (19,0.219854) (20,0.216561) (21,0.213598) (22,0.210852) (23,0.208561) (24,0.206426) (25,0.20438) (26,0.202266) (27,0.199495) (28,0.197597) (29,0.195185) (30,0.192938) (31,0.191131) (32,0.189076) (33,0.186821) (34,0.184871) (35,0.182511) (36,0.180476) (37,0.178677) (38,0.176755) (39,0.174803) (40,0.17334)
};
\label{plots:emp}

\end{axis}
\end{tikzpicture}&
%
%
\begin{tikzpicture}

\definecolor{mycolor1}{rgb}{0,0.4,0}
\definecolor{mycolor2}{rgb}{0,1,1}
\definecolor{mycolor3}{rgb}{1,0,1}
\definecolor{mycolor4}{rgb}{1,0.8,0.5}
\definecolor{mycolor5}{rgb}{0.7,0.4,0.01}

\begin{axis}[
footnotesize,
width= 1.6in,
height= 1.6in,
xmin=0, xmax=50,
ymin=0.5, ymax=1,
xlabel = {No. queried points},
ylabel = {Accuracy},
ytick={0.5,1},
xtick = {0,25,50},
axis on top,
axis y line = left,
axis x line = bottom
]
\addplot [
color=black,
solid,
line width=2.0pt
]
coordinates{ (1,0.562667) (2,0.753156) (3,0.793955) (4,0.908282) (5,0.953695) (6,0.960129) (7,0.966238) (8,0.971448) (9,0.973187) (10,0.976277) (11,0.978092) (12,0.978939) (13,0.980447) (14,0.981087) (15,0.981444) (16,0.982481) (17,0.982896) (18,0.983158) (19,0.9838) (20,0.983895) (21,0.984129) (22,0.984231) (23,0.98438) (24,0.984501) (25,0.984683) (26,0.984824) (27,0.984997) (28,0.985112) (29,0.985063) (30,0.985142) (31,0.985011) (32,0.98537) (33,0.985442) (34,0.985662) (35,0.985525) (36,0.985662) (37,0.985835) (38,0.985905) (39,0.986017) (40,0.98602) (41,0.986079) (42,0.986092) (43,0.986077) (44,0.986192) (45,0.986181) (46,0.986164) (47,0.986202) (48,0.986214) (49,0.986295) (50,0.986152)
};

\addplot [
color=red,
dashed,
line width=1.0pt
]
coordinates{ (1,0.567511) (2,0.583736) (3,0.602877) (4,0.605949) (5,0.604605) (6,0.631527) (7,0.632803) (8,0.662688) (9,0.682383) (10,0.716255) (11,0.716473) (12,0.717064) (13,0.722924) (14,0.80467) (15,0.810332) (16,0.811034) (17,0.838607) (18,0.848563) (19,0.852344) (20,0.852544) (21,0.852496) (22,0.874261) (23,0.87251) (24,0.873947) (25,0.874727) (26,0.873542) (27,0.874193) (28,0.87457) (29,0.885007) (30,0.901715) (31,0.906792) (32,0.913093) (33,0.913468) (34,0.918804) (35,0.91928) (36,0.92072) (37,0.92119) (38,0.922108) (39,0.923704) (40,0.923562) (41,0.924033) (42,0.924796) (43,0.92618) (44,0.92721) (45,0.926867) (46,0.926986) (47,0.930382) (48,0.930175) (49,0.931171) (50,0.931081)
};

\addplot [
color=green,
dashed,
line width=1.0pt
]
coordinates{ (1,0.570195) (2,0.790536) (3,0.769562) (4,0.792235) (5,0.793766) (6,0.789381) (7,0.793477) (8,0.794618) (9,0.799649) (10,0.799764) (11,0.800939) (12,0.840439) (13,0.840839) (14,0.962476) (15,0.969642) (16,0.976745) (17,0.978491) (18,0.978957) (19,0.980398) (20,0.980655) (21,0.981433) (22,0.981854) (23,0.982023) (24,0.982424) (25,0.982585) (26,0.982969) (27,0.983118) (28,0.983722) (29,0.984375) (30,0.984701) (31,0.984823) (32,0.984911) (33,0.985046) (34,0.985125) (35,0.985158) (36,0.985199) (37,0.985316) (38,0.985346) (39,0.985447) (40,0.985536) (41,0.985705) (42,0.985663) (43,0.985744) (44,0.98572) (45,0.985741) (46,0.985801) (47,0.98584) (48,0.985905) (49,0.985902) (50,0.985882)
};

\addplot [
color=mycolor1,
dashed,
line width=1.0pt
]
coordinates{ (1,0.5555) (2,0.752443) (3,0.808389) (4,0.914839) (5,0.952417) (6,0.960066) (7,0.96555) (8,0.970975) (9,0.972806) (10,0.975306) (11,0.977924) (12,0.978673) (13,0.980273) (14,0.980852) (15,0.981352) (16,0.982291) (17,0.982711) (18,0.98324) (19,0.983657) (20,0.983894) (21,0.984238) (22,0.98426) (23,0.984381) (24,0.984398) (25,0.984555) (26,0.984857) (27,0.984992) (28,0.984973) (29,0.985081) (30,0.985062) (31,0.985316) (32,0.985359) (33,0.985607) (34,0.985517) (35,0.985559) (36,0.985728) (37,0.985821) (38,0.985893) (39,0.986032) (40,0.986035) (41,0.985985) (42,0.986138) (43,0.986139) (44,0.986202) (45,0.986182) (46,0.986125) (47,0.98618) (48,0.986219) (49,0.986206) (50,0.986211)
};

\addplot [
color=mycolor2,
dashed,
line width=1.0pt
]
coordinates{ (1,0.567088) (2,0.756156) (3,0.77472) (4,0.84549) (5,0.888183) (6,0.935814) (7,0.950505) (8,0.953254) (9,0.96023) (10,0.961094) (11,0.962025) (12,0.962912) (13,0.963592) (14,0.963964) (15,0.9646) (16,0.964595) (17,0.96811) (18,0.968584) (19,0.969664) (20,0.971402) (21,0.971379) (22,0.971534) (23,0.971959) (24,0.972581) (25,0.973175) (26,0.973403) (27,0.974142) (28,0.974973) (29,0.97531) (30,0.975709) (31,0.976108) (32,0.976562) (33,0.977256) (34,0.978279) (35,0.978259) (36,0.978495) (37,0.978835) (38,0.979033) (39,0.978963) (40,0.979072) (41,0.979226) (42,0.979385) (43,0.979489) (44,0.979539) (45,0.979561) (46,0.979569) (47,0.979883) (48,0.979897) (49,0.980365) (50,0.980764)
};

\addplot [
color=blue,
dashed,
line width=1.0pt
]
coordinates{ (1,0.541424) (2,0.756255) (3,0.795493) (4,0.912543) (5,0.948411) (6,0.958323) (7,0.964496) (8,0.968224) (9,0.970146) (10,0.971164) (11,0.971896) (12,0.973024) (13,0.974036) (14,0.975054) (15,0.975318) (16,0.975498) (17,0.975662) (18,0.976209) (19,0.976408) (20,0.976869) (21,0.977163) (22,0.977486) (23,0.977661) (24,0.977664) (25,0.977685) (26,0.977809) (27,0.977947) (28,0.978116) (29,0.978213) (30,0.978352) (31,0.978303) (32,0.978754) (33,0.978882) (34,0.979323) (35,0.980083) (36,0.98021) (37,0.980292) (38,0.980424) (39,0.980634) (40,0.980771) (41,0.98108) (42,0.981038) (43,0.981107) (44,0.981237) (45,0.981289) (46,0.981433) (47,0.981373) (48,0.981827) (49,0.98176) (50,0.981863)
};

\addplot [
color=mycolor3,
dashed,
line width=1.0pt
]
coordinates{ (1,0.63282) (2,0.800935) (3,0.829853) (4,0.858889) (5,0.87209) (6,0.878661) (7,0.891389) (8,0.895449) (9,0.907277) (10,0.927994) (11,0.932065) (12,0.943548) (13,0.945201) (14,0.955266) (15,0.957909) (16,0.968668) (17,0.975885) (18,0.976961) (19,0.977784) (20,0.980258) (21,0.980665) (22,0.980936) (23,0.981566) (24,0.982013) (25,0.982599) (26,0.983531) (27,0.984044) (28,0.984459) (29,0.983794) (30,0.984794) (31,0.984868) (32,0.984892) (33,0.985008) (34,0.985095) (35,0.985132) (36,0.985239) (37,0.985262) (38,0.985339) (39,0.985378) (40,0.98544) (41,0.985478) (42,0.985559) (43,0.98561) (44,0.985618) (45,0.985677) (46,0.985665) (47,0.985771) (48,0.985762) (49,0.985863) (50,0.985847)
};

\addplot [
color=mycolor4,
dashed,
line width=1.0pt
]
coordinates{ (1,0.566305) (2,0.805511) (3,0.806916) (4,0.890281) (5,0.925929) (6,0.942951) (7,0.954453) (8,0.962239) (9,0.967025) (10,0.970886) (11,0.972673) (12,0.974604) (13,0.97687) (14,0.978388) (15,0.979488) (16,0.979832) (17,0.980456) (18,0.980774) (19,0.981082) (20,0.980673) (21,0.98151) (22,0.981714) (23,0.981879) (24,0.982011) (25,0.982005) (26,0.982162) (27,0.98215) (28,0.98217) (29,0.98237) (30,0.982652) (31,0.982849) (32,0.982782) (33,0.983037) (34,0.983002) (35,0.983031) (36,0.983396) (37,0.98363) (38,0.983661) (39,0.983856) (40,0.983978) (41,0.984125) (42,0.984153) (43,0.984425) (44,0.984461) (45,0.984743) (46,0.984859) (47,0.984994) (48,0.985047) (49,0.98508) (50,0.985148)
};

\addplot [
color=mycolor5,
dashed,
line width=1.0pt
]
coordinates{ (1,0.568585) (2,0.595898) (3,0.821141) (4,0.847498) (5,0.868668) (6,0.895947) (7,0.935528) (8,0.938798) (9,0.944039) (10,0.955205) (11,0.960002) (12,0.963834) (13,0.968172) (14,0.971887) (15,0.974662) (16,0.97714) (17,0.977679) (18,0.978224) (19,0.978781) (20,0.979608) (21,0.979789) (22,0.979978) (23,0.979955) (24,0.980202) (25,0.980297) (26,0.980368) (27,0.980489) (28,0.980387) (29,0.980479) (30,0.980839) (31,0.980877) (32,0.980866) (33,0.981077) (34,0.981198) (35,0.981205) (36,0.981136) (37,0.9812) (38,0.981187) (39,0.981212) (40,0.981213) (41,0.981259) (42,0.981259) (43,0.981284) (44,0.981398) (45,0.98127) (46,0.981704) (47,0.981701) (48,0.981791) (49,0.981785) (50,0.981759)
};

\addplot [
color=gray,
dashed,
line width=1.0pt
]
coordinates{ (1,0.563484) (2,0.565213) (3,0.566948) (4,0.565942) (5,0.564997) (6,0.565225) (7,0.563864) (8,0.563904) (9,0.563406) (10,0.564104) (11,0.55625) (12,0.551891) (13,0.596377) (14,0.601513) (15,0.604207) (16,0.604849) (17,0.605942) (18,0.606736) (19,0.60682) (20,0.607268) (21,0.607367) (22,0.608089) (23,0.607992) (24,0.608661) (25,0.608545) (26,0.608838) (27,0.60902) (28,0.609012) (29,0.609321) (30,0.609589) (31,0.609943) (32,0.610168) (33,0.610655) (34,0.610815) (35,0.610787) (36,0.611186) (37,0.611498) (38,0.61148) (39,0.611688) (40,0.611965) (41,0.611794) (42,0.612583) (43,0.612652) (44,0.612706) (45,0.612973) (46,0.612761) (47,0.613562) (48,0.613595) (49,0.613717) (50,0.613759)
};

\end{axis}
\end{tikzpicture}&
%
%
\begin{tikzpicture}

\definecolor{mycolor1}{rgb}{0,0.4,0}
\definecolor{mycolor2}{rgb}{0,1,1}
\definecolor{mycolor3}{rgb}{1,0,1}
\definecolor{mycolor4}{rgb}{1,0.8,0.5}
\definecolor{mycolor5}{rgb}{0.7,0.4,0.01}

\begin{axis}[
footnotesize,
width= 1.6in,
height= 1.6in,
xmin=0, xmax=50,
ymin=0.5, ymax=1,
ytick={0.5,1},
xtick = {0,25,50},
xlabel = {No. queried points},
ylabel = {Accuracy},
axis on top,
axis y line = left,
axis x line = bottom
]
\addplot [
color=black,
solid,
line width=2.0pt
]
coordinates{ (1,0.509885) (2,0.522693) (3,0.53474) (4,0.547013) (5,0.560323) (6,0.57036) (7,0.57857) (8,0.589043) (9,0.597232) (10,0.609322) (11,0.622262) (12,0.637122) (13,0.651602) (14,0.66623) (15,0.677149) (16,0.691247) (17,0.707813) (18,0.726329) (19,0.736819) (20,0.752558) (21,0.762286) (22,0.780063) (23,0.789513) (24,0.803286) (25,0.815501) (26,0.823959) (27,0.833154) (28,0.848929) (29,0.87118) (30,0.888573) (31,0.896232) (32,0.903101) (33,0.907358) (34,0.910533) (35,0.913087) (36,0.915929) (37,0.918698) (38,0.921288) (39,0.923818) (40,0.926381) (41,0.928526) (42,0.930591) (43,0.932818) (44,0.9349) (45,0.936965) (46,0.938718) (47,0.94052) (48,0.942503) (49,0.944208) (50,0.945905)
};

\addplot [
color=red,
dashed,
line width=1.0pt
]
coordinates{ (1,0.509773) (2,0.519229) (3,0.528577) (4,0.538652) (5,0.548848) (6,0.56459) (7,0.573944) (8,0.588079) (9,0.603113) (10,0.615583) (11,0.624322) (12,0.635052) (13,0.640151) (14,0.646008) (15,0.649586) (16,0.651762) (17,0.659059) (18,0.671112) (19,0.679088) (20,0.688642) (21,0.696339) (22,0.697418) (23,0.714628) (24,0.724863) (25,0.737618) (26,0.749363) (27,0.751161) (28,0.761194) (29,0.770765) (30,0.786319) (31,0.789465) (32,0.793527) (33,0.798071) (34,0.807288) (35,0.81155) (36,0.812729) (37,0.820657) (38,0.821573) (39,0.824516) (40,0.83211) (41,0.839197) (42,0.84107) (43,0.849292) (44,0.857284) (45,0.858584) (46,0.861788) (47,0.863781) (48,0.864775) (49,0.86646) (50,0.867783)
};

\addplot [
color=green,
dashed,
line width=1.0pt
]
coordinates{ (1,0.508449) (2,0.52448) (3,0.534489) (4,0.557607) (5,0.569195) (6,0.587917) (7,0.599197) (8,0.617996) (9,0.632287) (10,0.646611) (11,0.660148) (12,0.676873) (13,0.69087) (14,0.709665) (15,0.723865) (16,0.744137) (17,0.75402) (18,0.764207) (19,0.774987) (20,0.783602) (21,0.7938) (22,0.802727) (23,0.813786) (24,0.822881) (25,0.832758) (26,0.843949) (27,0.855273) (28,0.867317) (29,0.877196) (30,0.889835) (31,0.899085) (32,0.907538) (33,0.909411) (34,0.911749) (35,0.91423) (36,0.916021) (37,0.917998) (38,0.919709) (39,0.922173) (40,0.924) (41,0.925656) (42,0.927179) (43,0.930009) (44,0.93237) (45,0.933966) (46,0.935606) (47,0.937825) (48,0.940095) (49,0.941935) (50,0.943745)
};

\addplot [
color=mycolor1,
dashed,
line width=1.0pt
]
coordinates{ (1,0.508616) (2,0.515568) (3,0.525883) (4,0.537191) (5,0.547513) (6,0.554742) (7,0.564037) (8,0.573611) (9,0.584294) (10,0.594309) (11,0.60828) (12,0.622964) (13,0.634804) (14,0.647751) (15,0.661483) (16,0.67922) (17,0.689835) (18,0.703799) (19,0.716897) (20,0.729839) (21,0.74205) (22,0.752144) (23,0.768311) (24,0.778327) (25,0.786078) (26,0.798597) (27,0.814074) (28,0.833299) (29,0.848099) (30,0.868333) (31,0.89076) (32,0.901281) (33,0.906562) (34,0.90999) (35,0.913697) (36,0.916841) (37,0.919672) (38,0.922692) (39,0.925409) (40,0.928351) (41,0.930613) (42,0.9328) (43,0.934905) (44,0.937044) (45,0.939053) (46,0.940839) (47,0.942414) (48,0.944379) (49,0.94623) (50,0.947845)
};

\addplot [
color=mycolor2,
dashed,
line width=1.0pt
]
coordinates{ (1,0.511368) (2,0.521234) (3,0.534471) (4,0.541192) (5,0.549296) (6,0.55336) (7,0.566196) (8,0.576368) (9,0.590969) (10,0.604769) (11,0.616152) (12,0.631036) (13,0.644519) (14,0.655425) (15,0.671102) (16,0.681668) (17,0.699288) (18,0.719853) (19,0.731265) (20,0.738205) (21,0.746837) (22,0.750747) (23,0.772712) (24,0.783562) (25,0.798642) (26,0.80975) (27,0.812511) (28,0.821576) (29,0.835568) (30,0.840282) (31,0.8442) (32,0.849221) (33,0.853489) (34,0.86456) (35,0.869843) (36,0.872548) (37,0.87679) (38,0.88439) (39,0.889286) (40,0.892362) (41,0.896182) (42,0.900409) (43,0.902785) (44,0.905872) (45,0.908694) (46,0.911805) (47,0.915563) (48,0.918425) (49,0.921418) (50,0.923699)
};

\addplot [
color=blue,
dashed,
line width=1.0pt
]
coordinates{ (1,0.509498) (2,0.519054) (3,0.527929) (4,0.537568) (5,0.551151) (6,0.56285) (7,0.576111) (8,0.592502) (9,0.603575) (10,0.62071) (11,0.634414) (12,0.654218) (13,0.669335) (14,0.692337) (15,0.70947) (16,0.720645) (17,0.739182) (18,0.754014) (19,0.766625) (20,0.777609) (21,0.791608) (22,0.805876) (23,0.825049) (24,0.834503) (25,0.844029) (26,0.851438) (27,0.863511) (28,0.868529) (29,0.875978) (30,0.881048) (31,0.886339) (32,0.892711) (33,0.897363) (34,0.902574) (35,0.90609) (36,0.909593) (37,0.91294) (38,0.916156) (39,0.917961) (40,0.920041) (41,0.922319) (42,0.924864) (43,0.927053) (44,0.929225) (45,0.931453) (46,0.933538) (47,0.935713) (48,0.93771) (49,0.939482) (50,0.941397)
};

\addplot [
color=mycolor3,
dashed,
line width=1.0pt
]
coordinates{ (1,0.51337) (2,0.529771) (3,0.544297) (4,0.560177) (5,0.568214) (6,0.576879) (7,0.582433) (8,0.584435) (9,0.591358) (10,0.599803) (11,0.603702) (12,0.605604) (13,0.610153) (14,0.615944) (15,0.619902) (16,0.627108) (17,0.628215) (18,0.633332) (19,0.639372) (20,0.639883) (21,0.644345) (22,0.651738) (23,0.65781) (24,0.660388) (25,0.669717) (26,0.67488) (27,0.683468) (28,0.686149) (29,0.690429) (30,0.695878) (31,0.699595) (32,0.703835) (33,0.709444) (34,0.715295) (35,0.720755) (36,0.725257) (37,0.72849) (38,0.735388) (39,0.745585) (40,0.751332) (41,0.758135) (42,0.769001) (43,0.77831) (44,0.7876) (45,0.799152) (46,0.80651) (47,0.819522) (48,0.827645) (49,0.833988) (50,0.83782)
};

\addplot [
color=mycolor4,
dashed,
line width=1.0pt
]
coordinates{ (1,0.509503) (2,0.519416) (3,0.529021) (4,0.539572) (5,0.550976) (6,0.562576) (7,0.575101) (8,0.589434) (9,0.611601) (10,0.634933) (11,0.659819) (12,0.686802) (13,0.715388) (14,0.745879) (15,0.779305) (16,0.81661) (17,0.823452) (18,0.828173) (19,0.830868) (20,0.83343) (21,0.836034) (22,0.838609) (23,0.8412) (24,0.843833) (25,0.846406) (26,0.848906) (27,0.851831) (28,0.854557) (29,0.857288) (30,0.860204) (31,0.863002) (32,0.8658) (33,0.867865) (34,0.87011) (35,0.872043) (36,0.874039) (37,0.875788) (38,0.877721) (39,0.879712) (40,0.881449) (41,0.883304) (42,0.884793) (43,0.88689) (44,0.888692) (45,0.890531) (46,0.89233) (47,0.893886) (48,0.895479) (49,0.896751) (50,0.89818)
};

\addplot [
color=mycolor5,
dashed,
line width=1.0pt
]
coordinates{ (1,0.5108) (2,0.519728) (3,0.529965) (4,0.540986) (5,0.551291) (6,0.564799) (7,0.577927) (8,0.594231) (9,0.611973) (10,0.633352) (11,0.658299) (12,0.685346) (13,0.714151) (14,0.743465) (15,0.775744) (16,0.814025) (17,0.821312) (18,0.826681) (19,0.830442) (20,0.833942) (21,0.836694) (22,0.839855) (23,0.842425) (24,0.845011) (25,0.847563) (26,0.850202) (27,0.853042) (28,0.855901) (29,0.858893) (30,0.861653) (31,0.864464) (32,0.867183) (33,0.869036) (34,0.870872) (35,0.872713) (36,0.874727) (37,0.876214) (38,0.877867) (39,0.878996) (40,0.880543) (41,0.882084) (42,0.883558) (43,0.885244) (44,0.886777) (45,0.888309) (46,0.889937) (47,0.891268) (48,0.892703) (49,0.894114) (50,0.895325)
};

\addplot [
color=gray,
dashed,
line width=1.0pt
]
coordinates{ (1,0.509276) (2,0.519678) (3,0.520474) (4,0.520784) (5,0.521128) (6,0.520995) (7,0.521126) (8,0.521426) (9,0.521438) (10,0.521617) (11,0.521681) (12,0.521675) (13,0.521558) (14,0.521564) (15,0.521527) (16,0.521368) (17,0.521186) (18,0.520923) (19,0.520794) (20,0.520524) (21,0.520356) (22,0.520213) (23,0.520105) (24,0.519821) (25,0.519346) (26,0.519055) (27,0.520018) (28,0.520581) (29,0.520693) (30,0.521041) (31,0.521272) (32,0.521516) (33,0.521537) (34,0.521562) (35,0.521635) (36,0.521661) (37,0.521637) (38,0.521593) (39,0.521533) (40,0.521368) (41,0.521099) (42,0.52069) (43,0.520444) (44,0.520095) (45,0.519873) (46,0.519582) (47,0.519396) (48,0.518945) (49,0.518611) (50,0.518228)
};

\end{axis}
\end{tikzpicture} \\
\end{tabular}}
\caption{\emph{Top:} Evaluation on artificial datasets. Exemplars of the two classes are shown with black squares (\ref{plots:positives}) and red circles (\ref{plots:negatives}). \emph{Bottom:} Results of active learning with nine methods: random query (\ref{plots:rand}), \ourmethod (\ref{plots:BALD}),  MES (\ref{plots:maxent}), QBC with the vote criterion with 2 (\ref{plots:QBC2}) and 100 (\ref{plots:QBC100}) committee members, active SVM (\ref{plots:SVM}), IVM (\ref{plots:IVM}), decision theoretic: \cite{kapoor2007} (\ref{plots:dec}), \cite{zhu2003} (\ref{plots:semi}) and empirical error (\ref{plots:emp}).}
\label{fig:artificial}
\end{figure*}
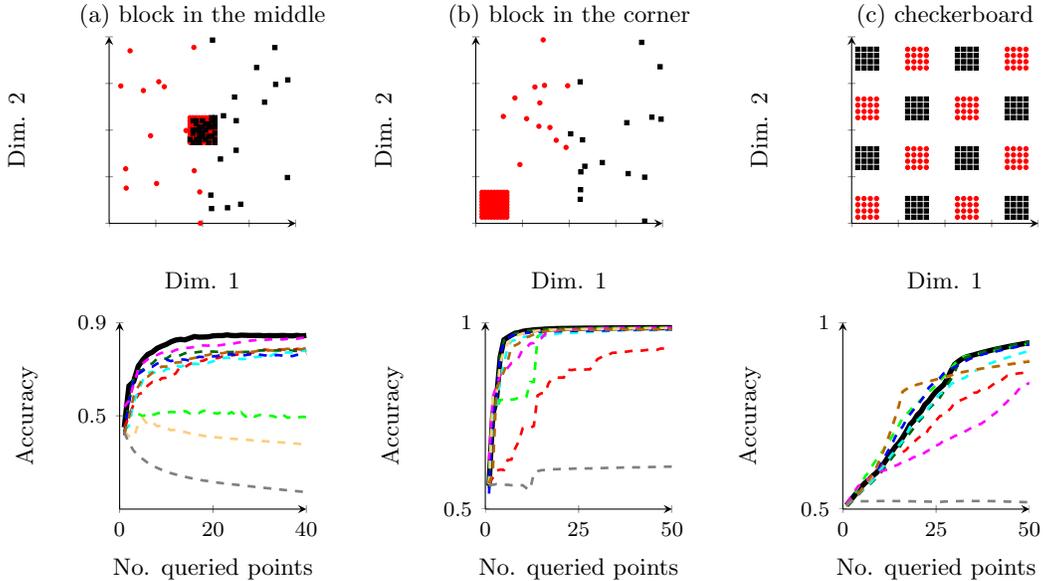

\begin{figure*}
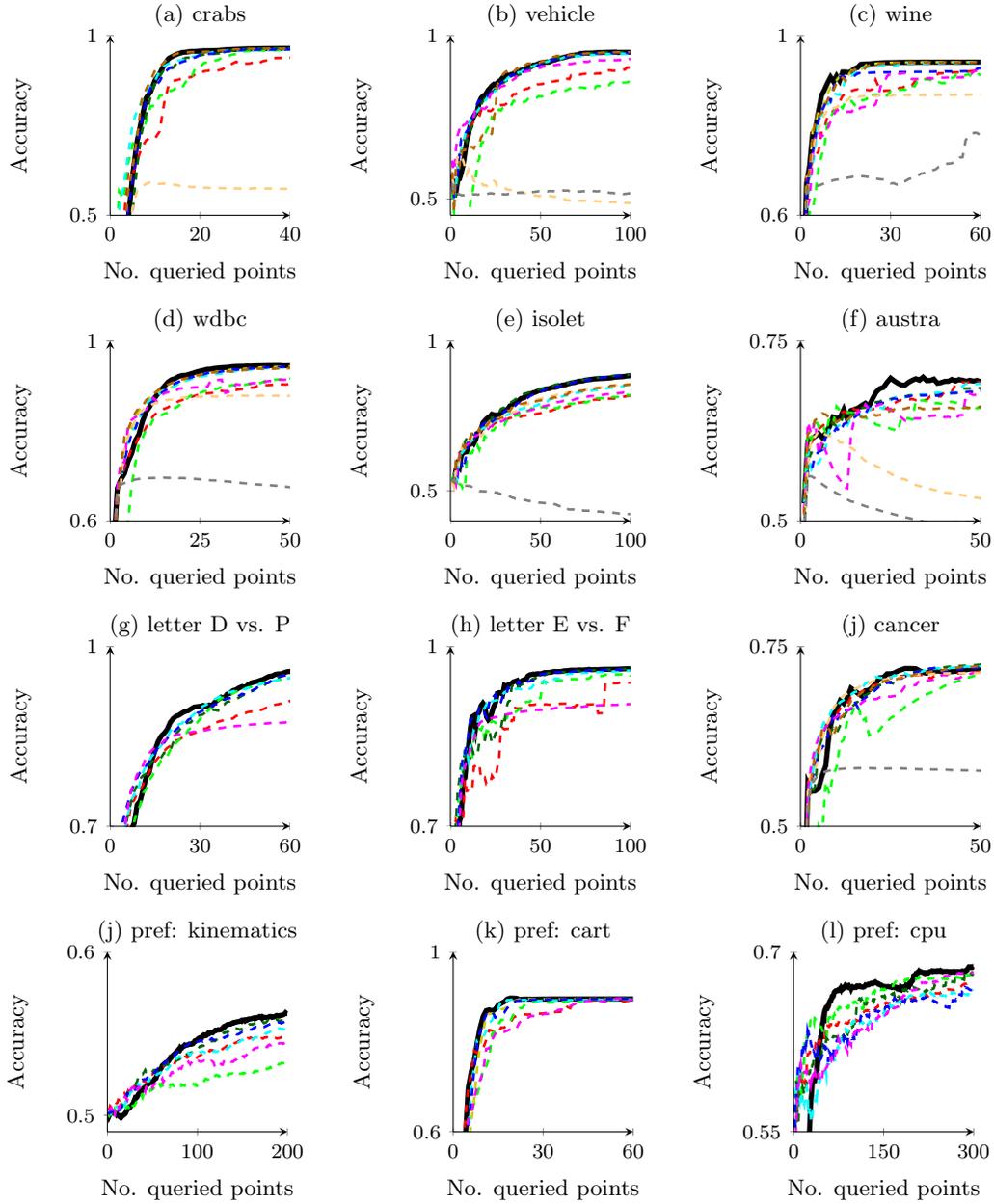

\noindent\makebox[\textwidth]{%
\begin{tabular}{ccc}
\input{crabs3.tikz}&
\input{vehicle2.tikz}&
%
%
\begin{tikzpicture}

\definecolor{mycolor1}{rgb}{0.8,0.8,0}
\definecolor{mycolor2}{rgb}{0,1,1}
\definecolor{mycolor3}{rgb}{1,0,1}
\definecolor{mycolor4}{rgb}{1,0.8,0.5}
\definecolor{mycolor5}{rgb}{0.7,0.4,0.01}

\begin{axis}[
footnotesize,
width= 1.6in,
height= 1.6in,
xmin=0, xmax=60,
ymin=0.6, ymax=1,
title={(c) wine},
xlabel = {No. queried points},
ylabel = {Accuracy},
ytick={0.6,1},
xtick = {0,30,60},
axis on top,
axis y line = left,
axis x line = bottom
]
\addplot [
color=black,
solid,
line width=2.0pt
]
coordinates{
 (1,0.464346)(2,0.67158)(3,0.710705)(4,0.782197)(5,0.794467)(6,0.833316)(7,0.866859)(8,0.879718)(9,0.894474)(10,0.907274)(11,0.897247)(12,0.909839)(13,0.899609)(14,0.91867)(15,0.918714)(16,0.917565)(17,0.920504)(18,0.926471)(19,0.928691)(20,0.935878)(21,0.936626)(22,0.93697)(23,0.937088)(24,0.937839)(25,0.938255)(26,0.938932)(27,0.939382)(28,0.939667)(29,0.939842)(30,0.939831)(31,0.939977)(32,0.9401)(33,0.940358)(34,0.940397)(35,0.940421)(36,0.940492)(37,0.940592)(38,0.940604)(39,0.940579)(40,0.94068)(41,0.940808)(42,0.940736)(43,0.940772)(44,0.94083)(45,0.94083)(46,0.94088)(47,0.940887)(48,0.940884)(49,0.940921)(50,0.94093)(51,0.940952)(52,0.940924)(53,0.940682)(54,0.940941)(55,0.940958)(56,0.940943)(57,0.940958)(58,0.940983)(59,0.940972)(60,0.940973) 
};

\addplot [
color=red,
dashed,
line width=1.0pt
]
coordinates{
 (1,0.436872)(2,0.684336)(3,0.696032)(4,0.724175)(5,0.733717)(6,0.742374)(7,0.787526)(8,0.794701)(9,0.82109)(10,0.824764)(11,0.832844)(12,0.835631)(13,0.852312)(14,0.865997)(15,0.869729)(16,0.870748)(17,0.876767)(18,0.879139)(19,0.880405)(20,0.878219)(21,0.878474)(22,0.879054)(23,0.878028)(24,0.880597)(25,0.901287)(26,0.886025)(27,0.87711)(28,0.877208)(29,0.877703)(30,0.877647)(31,0.878962)(32,0.87903)(33,0.883738)(34,0.883578)(35,0.889946)(36,0.8878)(37,0.888048)(38,0.897273)(39,0.898343)(40,0.898447)(41,0.902316)(42,0.902589)(43,0.902853)(44,0.911565)(45,0.909299)(46,0.916066)(47,0.915974)(48,0.916209)(49,0.916028)(50,0.917338)(51,0.918015)(52,0.918023)(53,0.918641)(54,0.920069)(55,0.920253)(56,0.920077)(57,0.920879)(58,0.92232)(59,0.92234)(60,0.916912) 
};

\addplot [
color=green,
dashed,
line width=1.0pt
]
coordinates{
 (1,0.423692)(2,0.520165)(3,0.607225)(4,0.610097)(5,0.650826)(6,0.708325)(7,0.732406)(8,0.75654)(9,0.806384)(10,0.819657)(11,0.829842)(12,0.839365)(13,0.845262)(14,0.838598)(15,0.854754)(16,0.85363)(17,0.854103)(18,0.84552)(19,0.855961)(20,0.855969)(21,0.856316)(22,0.858486)(23,0.871872)(24,0.867944)(25,0.868137)(26,0.868005)(27,0.862993)(28,0.861555)(29,0.868099)(30,0.875714)(31,0.87621)(32,0.87815)(33,0.879499)(34,0.879699)(35,0.879673)(36,0.887405)(37,0.887208)(38,0.887041)(39,0.899742)(40,0.900056)(41,0.901194)(42,0.903064)(43,0.90308)(44,0.896097)(45,0.896346)(46,0.896279)(47,0.894206)(48,0.894287)(49,0.904544)(50,0.905859)(51,0.906159)(52,0.913919)(53,0.91456)(54,0.911765)(55,0.911851)(56,0.913567)(57,0.90828)(58,0.912013)(59,0.915646)(60,0.910813) 
};

\addplot [
color=mycolor1,
dashed,
line width=1.0pt
]
coordinates{
 (1,0.445299)(2,0.685209)(3,0.687523)(4,0.788781)(5,0.802939)(6,0.833198)(7,0.833538)(8,0.850527)(9,0.869973)(10,0.884767)(11,0.893201)(12,0.90053)(13,0.909301)(14,0.908155)(15,0.908898)(16,0.912216)(17,0.922337)(18,0.921333)(19,0.924688)(20,0.92388)(21,0.928543)(22,0.934888)(23,0.936753)(24,0.936735)(25,0.937348)(26,0.938139)(27,0.937456)(28,0.939086)(29,0.939399)(30,0.939569)(31,0.939692)(32,0.939763)(33,0.940102)(34,0.940058)(35,0.940253)(36,0.940225)(37,0.940465)(38,0.94053)(39,0.940579)(40,0.940629)(41,0.940683)(42,0.940706)(43,0.94081)(44,0.940848)(45,0.940793)(46,0.940847)(47,0.940906)(48,0.940927)(49,0.94093)(50,0.940832)(51,0.940948)(52,0.940905)(53,0.940923)(54,0.940905)(55,0.940942)(56,0.940972)(57,0.940943)(58,0.940965)(59,0.940981)(60,0.941037) 
};

\addplot [
color=mycolor2,
dashed,
line width=1.0pt
]
coordinates{
 (1,0.480341)(2,0.582684)(3,0.680846)(4,0.709121)(5,0.760792)(6,0.817599)(7,0.822266)(8,0.840564)(9,0.846959)(10,0.85204)(11,0.87237)(12,0.869138)(13,0.867468)(14,0.880476)(15,0.903289)(16,0.91245)(17,0.910451)(18,0.917024)(19,0.917759)(20,0.918709)(21,0.929255)(22,0.929466)(23,0.930564)(24,0.930709)(25,0.932641)(26,0.932605)(27,0.932908)(28,0.932895)(29,0.933328)(30,0.931258)(31,0.931573)(32,0.932219)(33,0.937812)(34,0.938493)(35,0.938531)(36,0.938616)(37,0.938767)(38,0.938642)(39,0.938412)(40,0.938622)(41,0.938752)(42,0.939077)(43,0.939081)(44,0.939098)(45,0.939174)(46,0.93924)(47,0.939617)(48,0.939642)(49,0.93954)(50,0.939567)(51,0.939612)(52,0.939753)(53,0.939932)(54,0.939929)(55,0.940016)(56,0.94014)(57,0.940321)(58,0.940483)(59,0.940452)(60,0.940458) 
};

\addplot [
color=blue,
dashed,
line width=1.0pt
]
coordinates{
 (1,0.463948)(2,0.659312)(3,0.704106)(4,0.736769)(5,0.774687)(6,0.811565)(7,0.830289)(8,0.844535)(9,0.86365)(10,0.853837)(11,0.883514)(12,0.903617)(13,0.907196)(14,0.909918)(15,0.907093)(16,0.90763)(17,0.909226)(18,0.910732)(19,0.915457)(20,0.916963)(21,0.917427)(22,0.915942)(23,0.917169)(24,0.917439)(25,0.917729)(26,0.91795)(27,0.918038)(28,0.918097)(29,0.918035)(30,0.918323)(31,0.918554)(32,0.918555)(33,0.918524)(34,0.918633)(35,0.918761)(36,0.918732)(37,0.91898)(38,0.919033)(39,0.919079)(40,0.919095)(41,0.919259)(42,0.919311)(43,0.919476)(44,0.919513)(45,0.919751)(46,0.919816)(47,0.919919)(48,0.920294)(49,0.92)(50,0.920058)(51,0.920046)(52,0.920334)(53,0.92035)(54,0.920458)(55,0.920534)(56,0.920575)(57,0.920522)(58,0.920644)(59,0.92703)(60,0.927104) 
};

\addplot [
color=mycolor3,
dashed,
line width=1.0pt
]
coordinates{
 (1,0.485926)(2,0.612455)(3,0.63663)(4,0.727442)(5,0.782994)(6,0.784562)(7,0.775348)(8,0.780818)(9,0.791606)(10,0.808865)(11,0.823116)(12,0.833963)(13,0.833119)(14,0.832703)(15,0.829789)(16,0.830627)(17,0.83014)(18,0.838076)(19,0.83338)(20,0.839641)(21,0.840899)(22,0.842274)(23,0.842827)(24,0.842265)(25,0.847131)(26,0.867311)(27,0.895524)(28,0.898477)(29,0.911186)(30,0.913543)(31,0.913887)(32,0.914721)(33,0.914987)(34,0.916171)(35,0.916361)(36,0.912156)(37,0.903384)(38,0.903086)(39,0.903227)(40,0.903954)(41,0.904155)(42,0.904762)(43,0.9048)(44,0.90453)(45,0.904802)(46,0.905693)(47,0.906149)(48,0.906197)(49,0.906601)(50,0.906519)(51,0.906741)(52,0.906658)(53,0.906867)(54,0.906828)(55,0.912503)(56,0.912529)(57,0.915544)(58,0.92134)(59,0.921289)(60,0.921361) 
};

\addplot [
color=mycolor4,
dashed,
line width=1.0pt
]
coordinates{
 (1,0.48307)(2,0.624312)(3,0.700234)(4,0.738897)(5,0.776924)(6,0.790141)(7,0.813768)(8,0.824073)(9,0.834363)(10,0.841371)(11,0.838636)(12,0.839654)(13,0.847028)(14,0.852723)(15,0.853986)(16,0.857979)(17,0.866198)(18,0.867216)(19,0.863304)(20,0.864243)(21,0.865211)(22,0.864971)(23,0.864794)(24,0.865231)(25,0.865887)(26,0.866297)(27,0.866663)(28,0.866811)(29,0.866776)(30,0.866066)(31,0.86714)(32,0.867017)(33,0.867183)(34,0.867399)(35,0.867418)(36,0.867475)(37,0.86727)(38,0.867389)(39,0.867375)(40,0.867471)(41,0.867482)(42,0.867562)(43,0.867617)(44,0.867575)(45,0.867695)(46,0.867618)(47,0.86773)(48,0.867831)(49,0.8678)(50,0.867869)(51,0.867856)(52,0.867912)(53,0.867795)(54,0.868034)(55,0.868071)(56,0.868027)(57,0.868169)(58,0.867969)(59,0.868106)(60,0.868176) 
};

\addplot [
color=mycolor5,
dashed,
line width=1.0pt
]
coordinates{
 (1,0.448349)(2,0.647967)(3,0.708674)(4,0.775318)(5,0.808372)(6,0.820946)(7,0.82721)(8,0.837787)(9,0.856537)(10,0.873345)(11,0.878573)(12,0.894835)(13,0.900982)(14,0.906298)(15,0.899006)(16,0.914459)(17,0.920274)(18,0.924276)(19,0.922869)(20,0.922927)(21,0.923646)(22,0.931175)(23,0.934035)(24,0.935382)(25,0.936295)(26,0.936806)(27,0.937831)(28,0.938463)(29,0.938922)(30,0.939193)(31,0.939477)(32,0.939489)(33,0.939717)(34,0.939817)(35,0.93995)(36,0.940039)(37,0.940138)(38,0.940089)(39,0.9404)(40,0.939734)(41,0.940465)(42,0.940513)(43,0.940538)(44,0.940508)(45,0.940501)(46,0.940496)(47,0.94063)(48,0.940623)(49,0.941099)(50,0.940658)(51,0.940676)(52,0.940645)(53,0.94069)(54,0.940745)(55,0.940708)(56,0.940735)(57,0.940713)(58,0.940746)(59,0.940652)(60,0.940838) 
};

\addplot [
color=gray,
dashed,
line width=1.0pt
]
coordinates{
 (1,0.441817)(2,0.648168)(3,0.660785)(4,0.661994)(5,0.665923)(6,0.666059)(7,0.668773)(8,0.671377)(9,0.673379)(10,0.675978)(11,0.678188)(12,0.679489)(13,0.680056)(14,0.682275)(15,0.682444)(16,0.684333)(17,0.685221)(18,0.687453)(19,0.686741)(20,0.687062)(21,0.686852)(22,0.685867)(23,0.685131)(24,0.683866)(25,0.682224)(26,0.680735)(27,0.68001)(28,0.67986)(29,0.675824)(30,0.671218)(31,0.671579)(32,0.668688)(33,0.668918)(34,0.674676)(35,0.6764)(36,0.680726)(37,0.682593)(38,0.685932)(39,0.687362)(40,0.692615)(41,0.693699)(42,0.696056)(43,0.697926)(44,0.700153)(45,0.702384)(46,0.707544)(47,0.709034)(48,0.713244)(49,0.715006)(50,0.715286)(51,0.717213)(52,0.72091)(53,0.728328)(54,0.729374)(55,0.766766)(56,0.77701)(57,0.781245)(58,0.783873)(59,0.782871)(60,0.776154) 
};

\end{axis}
\end{tikzpicture}\\
%
%
\begin{tikzpicture}

\definecolor{mycolor1}{rgb}{0.8,0.8,0}
\definecolor{mycolor2}{rgb}{0,1,1}
\definecolor{mycolor3}{rgb}{1,0,1}
\definecolor{mycolor4}{rgb}{1,0.8,0.5}
\definecolor{mycolor5}{rgb}{0.7,0.4,0.01}

\begin{axis}[
footnotesize,
width= 1.6in,
height= 1.6in,
xmin=0, xmax=50,
ymin=0.6, ymax=1,
title={(d) wdbc},
xlabel = {No. queried points},
ylabel = {Accuracy},
ytick={0.6,1},
xtick = {0,25,50},
axis on top,
axis y line = left,
axis x line = bottom
]
\addplot [
color=black,
solid,
line width=2.0pt
]
coordinates{
 (1,0.538772)(2,0.672476)(3,0.691876)(4,0.705343)(5,0.732245)(6,0.750364)(7,0.77824)(8,0.792601)(9,0.821792)(10,0.843088)(11,0.854194)(12,0.869838)(13,0.879588)(14,0.889851)(15,0.898761)(16,0.908183)(17,0.91383)(18,0.916238)(19,0.920938)(20,0.923367)(21,0.923721)(22,0.925909)(23,0.928436)(24,0.931059)(25,0.933213)(26,0.934985)(27,0.937095)(28,0.937399)(29,0.939118)(30,0.939825)(31,0.94059)(32,0.941098)(33,0.941782)(34,0.942872)(35,0.943374)(36,0.94363)(37,0.943711)(38,0.944312)(39,0.944019)(40,0.944544)(41,0.944804)(42,0.944865)(43,0.945069)(44,0.945262)(45,0.945237)(46,0.94542)(47,0.945464)(48,0.94483)(49,0.944802)(50,0.945109) 
};

\addplot [
color=red,
dashed,
line width=1.0pt
]
coordinates{
 (1,0.542329)(2,0.66702)(3,0.687943)(4,0.718299)(5,0.739077)(6,0.761798)(7,0.76898)(8,0.78697)(9,0.791489)(10,0.808683)(11,0.819829)(12,0.833788)(13,0.839957)(14,0.842726)(15,0.849222)(16,0.852112)(17,0.85728)(18,0.860324)(19,0.866072)(20,0.868726)(21,0.872375)(22,0.873896)(23,0.876723)(24,0.879022)(25,0.882153)(26,0.882778)(27,0.883546)(28,0.884246)(29,0.885201)(30,0.889002)(31,0.889324)(32,0.887866)(33,0.889426)(34,0.891379)(35,0.891276)(36,0.892407)(37,0.893255)(38,0.894671)(39,0.896006)(40,0.896526)(41,0.900215)(42,0.90153)(43,0.90148)(44,0.902005)(45,0.902483)(46,0.903059)(47,0.903551)(48,0.903459)(49,0.903242)(50,0.905455) 
};

\addplot [
color=green,
dashed,
line width=1.0pt
]
coordinates{
 (1,0.501136)(2,0.569518)(3,0.5645)(4,0.580953)(5,0.598932)(6,0.680395)(7,0.727003)(8,0.772528)(9,0.784526)(10,0.803381)(11,0.808062)(12,0.826233)(13,0.83032)(14,0.833602)(15,0.83826)(16,0.84031)(17,0.846484)(18,0.850776)(19,0.855287)(20,0.864535)(21,0.869732)(22,0.876702)(23,0.880916)(24,0.883249)(25,0.884372)(26,0.887944)(27,0.889236)(28,0.892759)(29,0.892961)(30,0.894359)(31,0.898143)(32,0.899251)(33,0.900511)(34,0.901989)(35,0.902426)(36,0.901335)(37,0.901934)(38,0.90347)(39,0.904213)(40,0.905228)(41,0.905907)(42,0.907167)(43,0.910753)(44,0.912825)(45,0.913214)(46,0.914083)(47,0.914382)(48,0.9149)(49,0.915813)(50,0.916667) 
};

\addplot [
color=mycolor1,
dashed,
line width=1.0pt
]
coordinates{
 (1,0.535859)(2,0.653869)(3,0.689109)(4,0.740365)(5,0.786991)(6,0.8085)(7,0.819614)(8,0.846821)(9,0.854222)(10,0.865444)(11,0.874448)(12,0.882753)(13,0.887076)(14,0.887363)(15,0.891386)(16,0.893097)(17,0.896462)(18,0.902374)(19,0.909398)(20,0.912889)(21,0.912962)(22,0.917257)(23,0.918564)(24,0.920305)(25,0.922086)(26,0.923886)(27,0.925592)(28,0.928399)(29,0.930466)(30,0.931946)(31,0.932807)(32,0.934169)(33,0.935423)(34,0.936875)(35,0.93638)(36,0.93708)(37,0.937885)(38,0.938582)(39,0.939451)(40,0.940053)(41,0.940393)(42,0.941196)(43,0.941555)(44,0.941872)(45,0.941679)(46,0.942273)(47,0.942631)(48,0.943029)(49,0.943586)(50,0.943879) 
};

\addplot [
color=mycolor2,
dashed,
line width=1.0pt
]
coordinates{
 (1,0.534961)(2,0.661643)(3,0.695806)(4,0.732331)(5,0.77409)(6,0.801397)(7,0.806171)(8,0.82456)(9,0.843392)(10,0.860917)(11,0.86776)(12,0.8736)(13,0.878565)(14,0.88384)(15,0.890229)(16,0.891951)(17,0.895542)(18,0.899313)(19,0.905134)(20,0.910246)(21,0.911327)(22,0.913872)(23,0.91661)(24,0.917678)(25,0.920189)(26,0.922451)(27,0.922887)(28,0.923716)(29,0.925501)(30,0.927406)(31,0.927471)(32,0.929763)(33,0.931858)(34,0.932321)(35,0.933207)(36,0.93437)(37,0.934481)(38,0.935294)(39,0.93649)(40,0.936971)(41,0.938101)(42,0.938031)(43,0.939189)(44,0.939371)(45,0.939522)(46,0.939853)(47,0.940809)(48,0.940913)(49,0.941188)(50,0.941655) 
};

\addplot [
color=blue,
dashed,
line width=1.0pt
]
coordinates{
 (1,0.538022)(2,0.649796)(3,0.691373)(4,0.7347)(5,0.765972)(6,0.796137)(7,0.805401)(8,0.828459)(9,0.84375)(10,0.846641)(11,0.856468)(12,0.866783)(13,0.874249)(14,0.883381)(15,0.88813)(16,0.893767)(17,0.900802)(18,0.905443)(19,0.908868)(20,0.913212)(21,0.916944)(22,0.917857)(23,0.920065)(24,0.922748)(25,0.923896)(26,0.926072)(27,0.928376)(28,0.92995)(29,0.928792)(30,0.930871)(31,0.932411)(32,0.933139)(33,0.935135)(34,0.936851)(35,0.936208)(36,0.936621)(37,0.937515)(38,0.938344)(39,0.938702)(40,0.939122)(41,0.939821)(42,0.940145)(43,0.940856)(44,0.941383)(45,0.941511)(46,0.94179)(47,0.942136)(48,0.942466)(49,0.942837)(50,0.942961) 
};

\addplot [
color=mycolor3,
dashed,
line width=1.0pt
]
coordinates{
 (1,0.534085)(2,0.674385)(3,0.733316)(4,0.728263)(5,0.811499)(6,0.824549)(7,0.830668)(8,0.833809)(9,0.844859)(10,0.84781)(11,0.859001)(12,0.87301)(13,0.866695)(14,0.877437)(15,0.883526)(16,0.886431)(17,0.88431)(18,0.886878)(19,0.894221)(20,0.894706)(21,0.894103)(22,0.89482)(23,0.896726)(24,0.897127)(25,0.896886)(26,0.897171)(27,0.897486)(28,0.912312)(29,0.912667)(30,0.913357)(31,0.913817)(32,0.884484)(33,0.887062)(34,0.889077)(35,0.903975)(36,0.90285)(37,0.903172)(38,0.904714)(39,0.905628)(40,0.906445)(41,0.906771)(42,0.907327)(43,0.908938)(44,0.910873)(45,0.910939)(46,0.911307)(47,0.913921)(48,0.913803)(49,0.914132)(50,0.91856) 
};

\addplot [
color=mycolor4,
dashed,
line width=1.0pt
]
coordinates{
 (1,0.542286)(2,0.647676)(3,0.684299)(4,0.765408)(5,0.785661)(6,0.80368)(7,0.824575)(8,0.838748)(9,0.844158)(10,0.854877)(11,0.856518)(12,0.860142)(13,0.862646)(14,0.862114)(15,0.868292)(16,0.869078)(17,0.869349)(18,0.87041)(19,0.872184)(20,0.873819)(21,0.874943)(22,0.875591)(23,0.874662)(24,0.875537)(25,0.875934)(26,0.876413)(27,0.876408)(28,0.876492)(29,0.8765)(30,0.876787)(31,0.876361)(32,0.877184)(33,0.877825)(34,0.878248)(35,0.87812)(36,0.878261)(37,0.878225)(38,0.878059)(39,0.878157)(40,0.878085)(41,0.878083)(42,0.878232)(43,0.878141)(44,0.878077)(45,0.877981)(46,0.878081)(47,0.878018)(48,0.877874)(49,0.877813)(50,0.877693) 
};

\addplot [
color=mycolor5,
dashed,
line width=1.0pt
]
coordinates{
 (1,0.534678)(2,0.643008)(3,0.71299)(4,0.787483)(5,0.803805)(6,0.822892)(7,0.840349)(8,0.851281)(9,0.854918)(10,0.868557)(11,0.874392)(12,0.882156)(13,0.888328)(14,0.895311)(15,0.898714)(16,0.90279)(17,0.906127)(18,0.909722)(19,0.911886)(20,0.914977)(21,0.91686)(22,0.919351)(23,0.921414)(24,0.923726)(25,0.924549)(26,0.927106)(27,0.927876)(28,0.930012)(29,0.930504)(30,0.93136)(31,0.931779)(32,0.932377)(33,0.932395)(34,0.931513)(35,0.932564)(36,0.934066)(37,0.935013)(38,0.936192)(39,0.936693)(40,0.937727)(41,0.938355)(42,0.937763)(43,0.937595)(44,0.938288)(45,0.938037)(46,0.9384)(47,0.939151)(48,0.939373)(49,0.939082)(50,0.939397) 
};

\addplot [
color=gray,
dashed,
line width=1.0pt
]
coordinates{
 (1,0.541932)(2,0.670576)(3,0.680694)(4,0.684042)(5,0.687374)(6,0.690178)(7,0.691761)(8,0.693175)(9,0.694408)(10,0.695219)(11,0.695042)(12,0.695246)(13,0.695591)(14,0.695615)(15,0.696036)(16,0.69549)(17,0.695536)(18,0.694999)(19,0.694929)(20,0.695142)(21,0.694724)(22,0.69396)(23,0.693655)(24,0.692962)(25,0.692229)(26,0.691458)(27,0.690815)(28,0.689933)(29,0.689904)(30,0.68962)(31,0.688481)(32,0.687812)(33,0.687231)(34,0.686338)(35,0.685566)(36,0.684538)(37,0.683797)(38,0.682995)(39,0.682204)(40,0.681396)(41,0.680824)(42,0.680404)(43,0.679982)(44,0.679259)(45,0.678679)(46,0.678175)(47,0.677099)(48,0.676302)(49,0.675726)(50,0.674623) 
};

\end{axis}
\end{tikzpicture}&
\input{isolet2.tikz}&
\input{austra3.tikz}\\
\input{letterDP2.tikz}&
\input{letterEF2.tikz}&
\input{cancer3.tikz}\\
\input{prefkinem2.tikz}&
%
%
\begin{tikzpicture}

\definecolor{mycolor1}{rgb}{0.8,0.8,0}
\definecolor{mycolor2}{rgb}{0,1,1}
\definecolor{mycolor3}{rgb}{1,0,1}

\begin{axis}[
footnotesize,
width= 1.6in,
height= 1.6in,
xmin=0, xmax=60,
ymin=0.6, ymax=1,
ytick={0.6,1},
xtick = {0,30,60},
title = {(k) pref: cart},
xlabel = {No. queried points},
ylabel = {Accuracy},
axis on top,
axis y line = left,
axis x line = bottom
]
\addplot [
color=black,
solid,
line width=2.0pt
]
coordinates{
 (1,0.312526)(2,0.399089)(3,0.486865)(4,0.579879)(5,0.69681)(6,0.733167)(7,0.757932)(8,0.797963)(9,0.832337)(10,0.856839)(11,0.869297)(12,0.871057)(13,0.868473)(14,0.872422)(15,0.881547)(16,0.885553)(17,0.888968)(18,0.895191)(19,0.897405)(20,0.897236)(21,0.896926)(22,0.894437)(23,0.894701)(24,0.894961)(25,0.894942)(26,0.894788)(27,0.894786)(28,0.894947)(29,0.895015)(30,0.895085)(31,0.895123)(32,0.89521)(33,0.895329)(34,0.89532)(35,0.89531)(36,0.895289)(37,0.895277)(38,0.89528)(39,0.895307)(40,0.89534)(41,0.895347)(42,0.895338)(43,0.895357)(44,0.895359)(45,0.895381)(46,0.895366)(47,0.895381)(48,0.895387)(49,0.895391)(50,0.895385)(51,0.895391)(52,0.895398)(53,0.895418)(54,0.895409)(55,0.895405)(56,0.895414)(57,0.895421)(58,0.895423)(59,0.89544)(60,0.895439) 
};

\addplot [
color=red,
dashed,
line width=1.0pt
]
coordinates{
 (1,0.318594)(2,0.454105)(3,0.448128)(4,0.629858)(5,0.696363)(6,0.730729)(7,0.780531)(8,0.774608)(9,0.785957)(10,0.786159)(11,0.804273)(12,0.826371)(13,0.824866)(14,0.822412)(15,0.820237)(16,0.832047)(17,0.83179)(18,0.836967)(19,0.842231)(20,0.853653)(21,0.857299)(22,0.85763)(23,0.860293)(24,0.860948)(25,0.860402)(26,0.861371)(27,0.862224)(28,0.862451)(29,0.862677)(30,0.863027)(31,0.864018)(32,0.864104)(33,0.865001)(34,0.866294)(35,0.86655)(36,0.866856)(37,0.866033)(38,0.877621)(39,0.874609)(40,0.888817)(41,0.88988)(42,0.890965)(43,0.891523)(44,0.892792)(45,0.895276)(46,0.891347)(47,0.892214)(48,0.892906)(49,0.893137)(50,0.893938)(51,0.893955)(52,0.894953)(53,0.895039)(54,0.894872)(55,0.892761)(56,0.890195)(57,0.890143)(58,0.890884)(59,0.89084)(60,0.891411) 
};

\addplot [
color=green,
dashed,
line width=1.0pt
]
coordinates{
 (1,0.314429)(2,0.519589)(3,0.582216)(4,0.596253)(5,0.627407)(6,0.639079)(7,0.663746)(8,0.68303)(9,0.705061)(10,0.74308)(11,0.745618)(12,0.764147)(13,0.817605)(14,0.825806)(15,0.837967)(16,0.849089)(17,0.849161)(18,0.850037)(19,0.845547)(20,0.859748)(21,0.8699)(22,0.874523)(23,0.875227)(24,0.884002)(25,0.885614)(26,0.889135)(27,0.889284)(28,0.890326)(29,0.891655)(30,0.889256)(31,0.887418)(32,0.887455)(33,0.887816)(34,0.887795)(35,0.887952)(36,0.888032)(37,0.88897)(38,0.889161)(39,0.889474)(40,0.889852)(41,0.890089)(42,0.890104)(43,0.890775)(44,0.891705)(45,0.892625)(46,0.893231)(47,0.893674)(48,0.89367)(49,0.893728)(50,0.894377)(51,0.891545)(52,0.892787)(53,0.893869)(54,0.893865)(55,0.894056)(56,0.894168)(57,0.894374)(58,0.894363)(59,0.894377)(60,0.89438) 
};

\addplot [
color=mycolor1,
dashed,
line width=1.0pt
]
coordinates{
 (1,0.324707)(2,0.412855)(3,0.395838)(4,0.435822)(5,0.539119)(6,0.620875)(7,0.671788)(8,0.742941)(9,0.79331)(10,0.822499)(11,0.858454)(12,0.862089)(13,0.862766)(14,0.86239)(15,0.861638)(16,0.872147)(17,0.873293)(18,0.877948)(19,0.884744)(20,0.890142)(21,0.892158)(22,0.894494)(23,0.894957)(24,0.895161)(25,0.89526)(26,0.895203)(27,0.895088)(28,0.895038)(29,0.895117)(30,0.895106)(31,0.89525)(32,0.895282)(33,0.895248)(34,0.895253)(35,0.895246)(36,0.895249)(37,0.895238)(38,0.895278)(39,0.895309)(40,0.895341)(41,0.895351)(42,0.895357)(43,0.895353)(44,0.89537)(45,0.895378)(46,0.895381)(47,0.895384)(48,0.895377)(49,0.895401)(50,0.895391)(51,0.895416)(52,0.895418)(53,0.895427)(54,0.895426)(55,0.895423)(56,0.895427)(57,0.895427)(58,0.895454)(59,0.895449)(60,0.895431) 
};

\addplot [
color=mycolor2,
dashed,
line width=1.0pt
]
coordinates{
 (1,0.317895)(2,0.452266)(3,0.460426)(4,0.559226)(5,0.566749)(6,0.695896)(7,0.745702)(8,0.804036)(9,0.805)(10,0.811118)(11,0.824262)(12,0.841579)(13,0.848588)(14,0.865537)(15,0.875016)(16,0.890557)(17,0.885396)(18,0.89312)(19,0.893051)(20,0.894387)(21,0.893036)(22,0.892429)(23,0.893162)(24,0.893525)(25,0.889412)(26,0.88941)(27,0.890437)(28,0.891215)(29,0.89128)(30,0.891927)(31,0.891926)(32,0.891964)(33,0.892077)(34,0.892123)(35,0.892213)(36,0.892291)(37,0.892623)(38,0.892715)(39,0.892702)(40,0.892732)(41,0.892906)(42,0.893074)(43,0.893121)(44,0.89322)(45,0.893265)(46,0.893563)(47,0.893649)(48,0.893608)(49,0.893619)(50,0.89361)(51,0.893656)(52,0.893651)(53,0.893636)(54,0.89367)(55,0.893715)(56,0.893655)(57,0.893578)(58,0.893612)(59,0.893685)(60,0.893575) 
};

\addplot [
color=blue,
dashed,
line width=1.0pt
]
coordinates{
 (1,0.311974)(2,0.451957)(3,0.454156)(4,0.539139)(5,0.620098)(6,0.664446)(7,0.738385)(8,0.798257)(9,0.822168)(10,0.843899)(11,0.860374)(12,0.866939)(13,0.867899)(14,0.853941)(15,0.856873)(16,0.861457)(17,0.864885)(18,0.874945)(19,0.8812)(20,0.888296)(21,0.891029)(22,0.89181)(23,0.891914)(24,0.891977)(25,0.89272)(26,0.893769)(27,0.893801)(28,0.893858)(29,0.894056)(30,0.894209)(31,0.894122)(32,0.894254)(33,0.894271)(34,0.894315)(35,0.894379)(36,0.894346)(37,0.894483)(38,0.894475)(39,0.894466)(40,0.894606)(41,0.89464)(42,0.894695)(43,0.894732)(44,0.89468)(45,0.894829)(46,0.89486)(47,0.894806)(48,0.894816)(49,0.894855)(50,0.894896)(51,0.894864)(52,0.894879)(53,0.894913)(54,0.894883)(55,0.894932)(56,0.895155)(57,0.895188)(58,0.895274)(59,0.895322)(60,0.895302) 
};

\addplot [
color=mycolor3,
dashed,
line width=1.0pt
]
coordinates{
 (1,0.310611)(2,0.380621)(3,0.479078)(4,0.506739)(5,0.626741)(6,0.635375)(7,0.674457)(8,0.700596)(9,0.71543)(10,0.734344)(11,0.784333)(12,0.785848)(13,0.805382)(14,0.817896)(15,0.822801)(16,0.82142)(17,0.822106)(18,0.824867)(19,0.827648)(20,0.831234)(21,0.828634)(22,0.842633)(23,0.847631)(24,0.85339)(25,0.853753)(26,0.857768)(27,0.857251)(28,0.857465)(29,0.856833)(30,0.861675)(31,0.862039)(32,0.862276)(33,0.8612)(34,0.866245)(35,0.870835)(36,0.871539)(37,0.871303)(38,0.876641)(39,0.879325)(40,0.882638)(41,0.885267)(42,0.887471)(43,0.888339)(44,0.890488)(45,0.890942)(46,0.890981)(47,0.89067)(48,0.891385)(49,0.890952)(50,0.88962)(51,0.889773)(52,0.892344)(53,0.893951)(54,0.895603)(55,0.896146)(56,0.896157)(57,0.896151)(58,0.893151)(59,0.89319)(60,0.893367) 
};

\end{axis}
\end{tikzpicture}&
\input{prefcpu2.tikz}
\end{tabular}}
\caption{Test set classification accuracy on classification and preference learning datasets. Methods used are \ourmethod (\ref{plots:BALD}), random query (\ref{plots:rand}), MES (\ref{plots:maxent}), QBC with 2 ($\mbox{QBC}_2$, \ref{plots:QBC2}) and 100 ($\mbox{QBC}_{100}$, \ref{plots:QBC100}) committee members, active SVM (\ref{plots:SVM}), IVM (\ref{plots:IVM}), decision theoretic \cite{kapoor2007} (\ref{plots:dec}), decision theoretic \cite{zhu2003} (\ref{plots:semi}) and empicial error (\ref{plots:emp}). The decision theoretic methods took a long time to run, so were not completed for all datasets. Plots (a-i) are GPC datasets, (j-l) are preference learning.}.
\label{fig:all}
\end{figure*}

\begin{figure*}
\noindent\makebox[\textwidth]{%
%
%
\begin{tikzpicture}

\begin{axis}[%
scale only axis,
width=4.5in,
height=1.8in,
xmin=0.5, xmax=9.5,
ymin=-10, ymax=120,
xticklabels={\empty},
axis on top]
\addplot [
color=black,
dashed
]
coordinates{
 (1,39.875)(1,57) 
};

\addplot [
color=black,
dashed
]
coordinates{
 (2,37.9423)(2,51.6) 
};

\addplot [
color=black,
dashed
]
coordinates{
 (3,4.0625)(3,5.05) 
};

\addplot [
color=black,
dashed
]
coordinates{
 (4,14.5673)(4,18) 
};

\addplot [
color=black,
dashed
]
coordinates{
 (5,4.05)(5,4.1) 
};

\addplot [
color=black,
dashed
]
coordinates{
 (6,38.8308)(6,57) 
};

\addplot [
color=black,
dashed
]
coordinates{
 (7,55.5)(7,55.5) 
};

\addplot [
color=black,
dashed
]
coordinates{
 (8,3.2)(8,3.2) 
};

\addplot [
color=black,
dashed
]
coordinates{
 (9,84)(9,115.231) 
};

\addplot [
color=black,
dashed
]
coordinates{
 (1,0)(1,12.125) 
};

\addplot [
color=black,
dashed
]
coordinates{
 (2,0)(2,17.05) 
};

\addplot [
color=black,
dashed
]
coordinates{
 (3,-0.8)(3,0.15) 
};

\addplot [
color=black,
dashed
]
coordinates{
 (4,-0.8)(4,1.65) 
};

\addplot [
color=black,
dashed
]
coordinates{
 (5,0)(5,1.75) 
};

\addplot [
color=black,
dashed
]
coordinates{
 (6,0)(6,10.75) 
};

\addplot [
color=black,
dashed
]
coordinates{
 (7,0)(7,29.85) 
};

\addplot [
color=black,
dashed
]
coordinates{
 (8,-5.83333)(8,-2) 
};

\addplot [
color=black,
dashed
]
coordinates{
 (9,0)(9,32.9) 
};

\addplot [
color=black,
solid
]
coordinates{
 (0.875,57)(1.125,57) 
};

\addplot [
color=black,
solid
]
coordinates{
 (1.875,51.6)(2.125,51.6) 
};

\addplot [
color=black,
solid
]
coordinates{
 (2.875,5.05)(3.125,5.05) 
};

\addplot [
color=black,
solid
]
coordinates{
 (3.875,18)(4.125,18) 
};

\addplot [
color=black,
solid
]
coordinates{
 (4.875,4.1)(5.125,4.1) 
};

\addplot [
color=black,
solid
]
coordinates{
 (5.875,57)(6.125,57) 
};

\addplot [
color=black,
solid
]
coordinates{
 (6.875,55.5)(7.125,55.5) 
};

\addplot [
color=black,
solid
]
coordinates{
 (7.875,3.2)(8.125,3.2) 
};

\addplot [
color=black,
solid
]
coordinates{
 (8.875,115.231)(9.125,115.231) 
};

\addplot [
color=black,
solid
]
coordinates{
 (0.875,0)(1.125,0) 
};

\addplot [
color=black,
solid
]
coordinates{
 (1.875,0)(2.125,0) 
};

\addplot [
color=black,
solid
]
coordinates{
 (2.875,-0.8)(3.125,-0.8) 
};

\addplot [
color=black,
solid
]
coordinates{
 (3.875,-0.8)(4.125,-0.8) 
};

\addplot [
color=black,
solid
]
coordinates{
 (4.875,0)(5.125,0) 
};

\addplot [
color=black,
solid
]
coordinates{
 (5.875,0)(6.125,0) 
};

\addplot [
color=black,
solid
]
coordinates{
 (6.875,0)(7.125,0) 
};

\addplot [
color=black,
solid
]
coordinates{
 (7.875,-5.83333)(8.125,-5.83333) 
};

\addplot [
color=black,
solid
]
coordinates{
 (8.875,0)(9.125,0) 
};

\addplot [
color=blue,
solid
]
coordinates{
 (0.75,12.125)(0.75,39.875)(1.25,39.875)(1.25,12.125)(0.75,12.125) 
};

\addplot [
color=blue,
solid
]
coordinates{
 (1.75,17.05)(1.75,37.9423)(2.25,37.9423)(2.25,17.05)(1.75,17.05) 
};

\addplot [
color=blue,
solid
]
coordinates{
 (2.75,0.15)(2.75,4.0625)(3.25,4.0625)(3.25,0.15)(2.75,0.15) 
};

\addplot [
color=blue,
solid
]
coordinates{
 (3.75,1.65)(3.75,14.5673)(4.25,14.5673)(4.25,1.65)(3.75,1.65) 
};

\addplot [
color=blue,
solid
]
coordinates{
 (4.75,1.75)(4.75,4.05)(5.25,4.05)(5.25,1.75)(4.75,1.75) 
};

\addplot [
color=blue,
solid
]
coordinates{
 (5.75,10.75)(5.75,38.8308)(6.25,38.8308)(6.25,10.75)(5.75,10.75) 
};

\addplot [
color=blue,
solid
]
coordinates{
 (6.75,29.85)(6.75,55.5)(7.25,55.5)(7.25,29.85)(6.75,29.85) 
};

\addplot [
color=blue,
solid
]
coordinates{
 (7.75,-2)(7.75,3.2)(8.25,3.2)(8.25,-2)(7.75,-2) 
};

\addplot [
color=blue,
solid
]
coordinates{
 (8.75,32.9)(8.75,84)(9.25,84)(9.25,32.9)(8.75,32.9) 
};

\addplot [
color=red,
solid
]
coordinates{
 (0.75,27.3)(1.25,27.3) 
};

\addplot [
color=red,
solid
]
coordinates{
 (1.75,23.8)(2.25,23.8) 
};

\addplot [
color=red,
solid
]
coordinates{
 (2.75,2.3)(3.25,2.3) 
};

\addplot [
color=red,
solid
]
coordinates{
 (3.75,3.95)(4.25,3.95) 
};

\addplot [
color=red,
solid
]
coordinates{
 (4.75,3.25)(5.25,3.25) 
};

\addplot [
color=red,
solid
]
coordinates{
 (5.75,16.2)(6.25,16.2) 
};

\addplot [
color=red,
solid
]
coordinates{
 (6.75,43.7)(7.25,43.7) 
};

\addplot [
color=red,
solid
]
coordinates{
 (7.75,1.35)(8.25,1.35) 
};

\addplot [
color=red,
solid
]
coordinates{
 (8.75,46.95)(9.25,46.95) 
};

\addplot [
color=blue,
only marks,
mark=+,
mark options={solid,draw=red}
]
coordinates{
 (5,19.8462) 
};

\addplot [
color=blue,
only marks,
mark=+,
mark options={solid,draw=red}
]
coordinates{
 (7,115.231) 
};

\addplot [
color=blue,
only marks,
mark=+,
mark options={solid,draw=red}
]
coordinates{
 (8,77.5385) 
};

\addplot [
color=red,
dashed
]
coordinates{
 (-0.5,0)(9.9,0) 
};

\end{axis}
\node[above, inner sep=0mm, text=black]
at (12, 0.2) {{BALD}};
\node[above, inner sep=0mm, text=black]
at (0.6, -1.6) {\rotatebox{90}{Rand}};
\node[above, inner sep=0mm, text=black]
at (1.8, -1.6) {\rotatebox{90}{IVM}};
\node[above, inner sep=0mm, text=black]
at (3.1, -1.6) {\rotatebox{90}{MES}};
\node[above, inner sep=0mm, text=black]
at (4.4, -1.6) {\rotatebox{90}{QBC2}};
\node[above, inner sep=0mm, text=black]
at (5.7, -1.6) {\rotatebox{90}{QBC100}};
\node[above, inner sep=0mm, text=black]
at (7.0, -1.6) {\rotatebox{90}{SVM}};
\node[above, inner sep=0mm, text=black]
at (8.3, -1.6) {\rotatebox{90}{Kapoor}};
\node[above, inner sep=0mm, text=black]
at (9.6, -1.6) {\rotatebox{90}{Zhu \emph{et al.}}};
\node[above, inner sep=0mm, text=black]
at (10.9, -1.6) {\rotatebox{90}{Empirical}};
\end{tikzpicture}
}
\caption{Summary of results for all classification experiments. $y$-axis denotes the number of additional data points, relative to BALD, required to achieve at least $97.5\%$ of the predictive performance of the entire pool. The `box' denotes 25th to 75th percentile, the red line denotes the median over datasets, and the `whiskers' depict the range. The crosses denote outliers ($>2.7\sigma$ from the mean). Positive values mean that the algorithm required more data points than BALD to achieve the same performance.}
\label{fig:boxandwhisker}
\end{figure*}
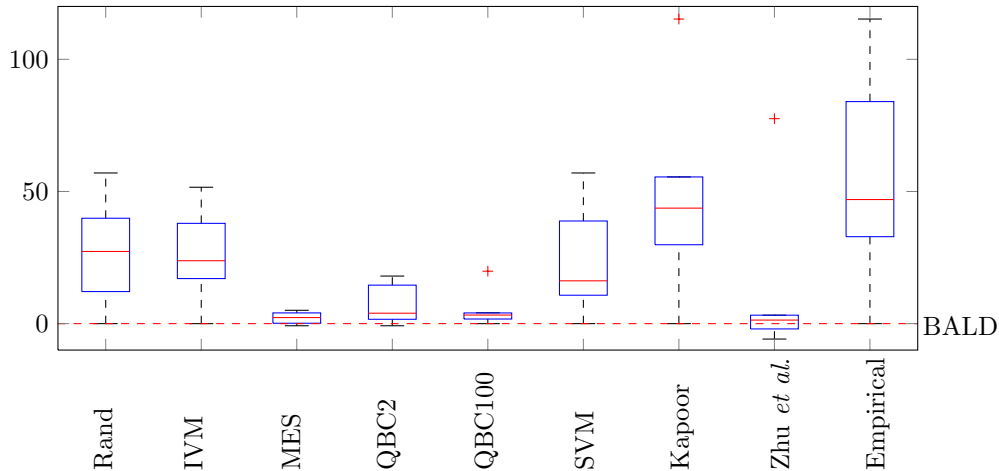

\paragraph{Quantifying Approximation Losses:} To obtain \eqref{eqn:BALD_GPC} we made two approximations: we perform approximate inference ({\scriptsize $\stackrel{1}{\approx}$}), and we approximated the binary entropy of the Gaussian CDF by a squared exponential ({\scriptsize $\stackrel{2}{\approx}$}). Both of these can be substituted with Monte Carlo sampling, enabling us to compute an asymptotically unbiased estimate of the expected information gain. Using extensive Monte Carlo as the `gold standard', we can evaluate how much we loose by applying these approximations. We quantify approximation error as: 

\begin{align}
\frac{ \max_{\x\in\mathcal{P}} I(\x) - I(\argmax_{\x\in\mathcal{P}}\hat{I}(\x)) }{{\max_{\x\in\mathcal{P}}I(\x) }}\cdot 100\% 
\end{align}

where $I$ is the objective computed using Monte Carlo, $\hat{I}$ is the approximate objective. The \emph{cancer} UCI dataset was used, results and discussion are in Fig.\,\ref{fig:trick_results}.

\paragraph{Pool based active learning:} We test \ourmethod for GPC and preference learning in the pool-based setting i.e. selecting $\x$ values from a fixed set of data-points. Although BALD can generalise to selecting continuous $\x$, this enables us to compare to algorithms that cannot. We compare to eight other algorithms: random sampling, MES, QBC (with 2 and 100 committee members), SVM with version space approximation \cite{tong2001}, decision theoretic approaches in \cite{kapoor2007, zhu2003} and directly minimizing expected empirical error (the last is not a widely used method, but is included for analysis of \cite{kapoor2007}).

We consider three artificial, but challenging, datasets. The first of which, \emph{block in the middle}, has a block of noisy points on the decision boundary, the second \emph{block in the corner}, has a block of uninformative points far from the decision boundary: a strong active learning algorithm should avoid these uninformative regions. The third is similar to the \emph{checkerboard} dataset in \cite{zhu2003}, and is designed to test the algorithm's capabilities to find multiple disjoint islands of points from one class. The three datasets and results using each algorithm are depicted in Fig.\,\ref{fig:artificial}.

Results are also presented on eight UCI classification datasets \emph{australia, crabs, vehicle, isolet, cancer, wine, wdbc} and \emph{letter}. \emph{Letter} is a multiclass dataset for which we select hard-to-distinguish letters E vs. F and D vs. P. For preference learning we use the \emph{cpu, cart} and \emph{kinematics} regression datasets \footnote{http://www.liacc.up.pt/~ltorgo/Regression/DataSets.html} processed to yield a preference task as described in \cite{chu2005}. Results are plotted in Fig.\,\ref{fig:all}, and Fig.\,\ref{fig:boxandwhisker} depicts an aggregation of the results.

\paragraph{Discussion:} Figs.\,\ref{fig:artificial} and \ref{fig:all} show that by using \ourmethod we make significant gains over naive random sampling in both the classification and preference learning domains. Relative to other active learning algorithms \ourmethod is consistently the best, or amongst the best performing algorithms on all datasets. On any individual dataset BALD's performance is often matched because we compare to many methods, and the more approximate algorithms can have good performance under different conditions. Fig.\,\ref{fig:boxandwhisker} reveals that BALD has the best overall performance; on average, all other methods require more data points to achieve the same classification accuracy. Zhu \emph{et al.}'s decision theoretic approach is closest, the median increase in the number of data points required is $1.4$ and zero (i.e. equivalent to BALD) is within the inter-quartile range. This algorithm, however, requires much more computational time and has access to the full set of test inputs, which BALD does not have. MES and QBC appear close in performance to BALD, but the zero line falls outside both of their inter-quartile ranges.

As expected, MES performs poorly on the noisy dataset (Fig.\,\ref{fig:artificial}(a)) because it discards knowledge of observation noise. When there is zero observation noise it is equivalent to BALD e.g. Fig.\,\ref{fig:artificial}(c). On many of the real-world datasets MES performs as well as BALD e.g. Fig.\,\ref{fig:all}(b, e), indicating that these datasets are mostly noise-free.

The IVM performs well on Fig.\,\ref{fig:artificial}(c), but pathologically on \ref{fig:artificial}(a); this is due to the fact that it biases selection towards points from only one class in the noisy cluster, reducing the posterior entropy rapidly but artificially. However, it also performs significantly worse than BALD on noise-free (indicated by MES's strong performance) datasets e.g. Fig.\,\ref{fig:all}(b). This implies that the IVM's posterior approximation or the ADF update are detrimental to the algorithm's performance.

QBC often yields only a small decrement in performance, the sampling approximation is often not too detrimental. However, it performs poorly on the noisy artificial dataset (Fig.\,\ref{fig:artificial}(a)) because the vote criterion is not maintaining a notion of inherent uncertainty, like MES. The SVM-based approach exhibits variable performance (it does well on Fig.\,\ref{fig:all}(d), but very poorly on \ref{fig:all}(f)). The performance is greatly effected by the approximation used, for consistency we present here one that yielded the most consistent good performance.

Decision theoretic approaches sometimes perform well, on \ref{fig:artificial}(c) they choose the first 16 points from the centre of each cluster as they are influenced by the surrounding unlabelled points. \ourmethod does not observe the unlabelled points so may not pick points from the centres. Fig.\,\ref{fig:boxandwhisker} reveals that BALD is performing as well as the method in \cite{zhu2003}, and outperforms the approach in \cite{kapoor2007}, despite not having access to the locations of the test points and having a significantly lower computational cost. The objective in \cite{kapoor2007} can fail, this is because one term in their objective function is the empirical error. The weight given to this term is determined by the relative sizes of the training and test set (and the associated losses). Directly minimizing empirical error usually performs very pathologically, picking only `safe' points. When the method in \cite{kapoor2007} assigns too much weight to this term, it can fail also.

Finally we note that BALD may occasionally perform poorly on the first few data points (e.g. Fig.\,\ref{fig:all}(l)). This is may be because the hyperparameters are fixed throughout the experiments to provide a fair comparison to algorithms incapable of incorporating hyperparameter learning. This may mean that given little data the GP model overfits, leading to BALD selecting abnormal query locations. Maintaining a distribution over hyperparameters can be done using MCMC, although this significantly increases computational time. Designing a general method to do this efficiently is a subject of further work. In practice, a simple heuristic such as picking the first few points randomly, and optimising hyperparameters will usually suffice.

\section{Conclusions}

We have demonstrated a method that applies the full information theoretic active learning criterion to GP classification that makes, as far as the authors are aware, the smallest number of approximations to date, and has as good computational complexity. We extend the GPC model to develop a new preference learning kernel, which enables us to apply our active learning algorithm directly to this domain also. The method can handle naturally active learning of kernel hyperparameters, which is a hard, mostly unsolved problem, for example in SVM active learning. One notable feature of our approach is that it is agnostic to the approximate inference methods used. This allows us to choose from a whole range of approximate inference methods, including EP, the Laplace approximation, ADF or even sparse online learning, and thereby make the trade off between computational complexity and accuracy. Our experimental performance compares favourably to many other active learning methods for classification, and even decision theoretic methods that have access to the test data and require much greater computational time.

{
\bibliographystyle{apalike}
\bibliography{bibliog}
}

\newpage
\section*{APPENDIX -- SUPPLEMENTARY MATERIAL}

\subsection*{Taylor Expansion for Approximation  $\stackrel{2}{\approx}$}

We perform a Taylor expansion on $ \ln \rmH[\Phi(x)]$ as follows:

\begin{align}
f(x) &= f(0) + \frac{f'(0)x}{1!} + \frac{f''(0)x^2}{2!} + \dots \nonumber \\
f(x) &= \ln \rmH[\Phi(x)] \nonumber \\
f'(x) &= -\frac{1}{\ln 2}\frac{\Phi'(x)}{\rmH[\Phi(x)]}\left[\ln\Phi(x) - \ln(1-\Phi(x))  \right] \nonumber \\ 
f''(x) &= \frac{1}{\ln 2}\frac{\Phi'(x)^2}{\rmH[\Phi(x)]^2}\left[\ln\Phi(x) - \ln(1-\Phi(x))  \right]\nonumber\\
& \qquad - \frac{1}{\ln 2}\frac{\Phi''(x)}{\rmH[\Phi(x)]}\left[\ln\Phi(x) - \ln(1-\Phi(x))  \right] \nonumber\\
& \qquad - \frac{1}{\ln 2}\frac{\Phi'(x)^2}{\rmH[\Phi(x)]}\left[\frac{1}{\Phi(x)} + \frac{1}{(1-\Phi(x)})  \right] \nonumber\\
\therefore \ln \rmH[\Phi(x)] \nonumber  &= 1 - \frac{1}{\pi\ln2}x^2 + \mathcal{O}(x^4)
\end{align}

Because the function is even, we can inspect that the $x^3$ term will be zero. Therefore, exponentiating, we make the approximation up to $\mathcal{O}(x^4)$: $$\rmH[\Phi(x)]\stackrel{2}{\approx}\exp\left({-\frac{x^2}{\pi\ln2}}\right) $$

\subsection*{Preference Kernel}

 The mean $\mu_{\mathrm{pref}}$, and covariance function $k_{\mathrm{pref}}$ of the GP over $g$ can be computed from the mean and covariance of $f\sim \mathrm{GP}(\mu,k)$ as follows:
\begin{align}
	k_{\mathrm{pref}}&([\bm{u}_i,\bm{v}_i],[\bm{u}_j,\bm{v}_j]) = Cov[g(\bm{u}_i,\bm{v}_i),g(\bm{u}_j,\bm{v}_j)]\notag\\
		&= Cov\left[\left(f(\bm{u}_i) - f(\bm{v}_i)\right) , \left(f(\bm{u}_i)  - f(\bm{v}_i)\right)\right]\notag\\
		&= \mathbb{E}\left[\left(f(\bm{u}_i) - f(\bm{v}_i)\right)\cdot \left(f(\bm{u}_i)  - f(\bm{v}_i)\right)\right]\notag\\
		& \qquad - \left(\mu(\bm{u}_i) -  \mu(\bm{v}_i)\right) \left(\mu(\bm{v}_j) - \mu(\bm{u}_i)\right)\notag\\
		&= k(\bm{u}_i,\bm{u}_j) + k(\bm{v}_i,\bm{v}_j) \notag \\
		&\qquad - k(\bm{u}_i,\bm{v}_j) - k(\bm{v}_i,\bm{u}_j)\\
	\mu_{\mathrm{pref}}&([\bm{u},\bm{v}]) = \mathbb{E}\left[g([\bm{u},\bm{v}])\right] = \mathbb{E}\left[f(\bm{u}) - f(\bm{v})\right]\notag\\
		&=\mu(\bm{u}) - \mu(\bm{v})
\end{align}

\end{document}